\documentclass[10pt]{article}
\usepackage{amsfonts,amsmath,amsthm,amssymb}
\usepackage{graphicx}
\usepackage{fullpage}
\usepackage{float}
\usepackage{algorithm,algpseudocode}
\usepackage{hyperref}

\usepackage{diagbox}

\usepackage{xargs}
\usepackage[textsize=scriptsize]{todonotes}\reversemarginpar
\newcommandx{\emreNote}[2][1=]{\todo[inline,linecolor=black,backgroundcolor=green!25,bordercolor=green,#1]{#2 ---Emre}}

\newcommandx{\gidNote}[2][1=]{\todo[inline,linecolor=black,backgroundcolor=red!25,bordercolor=red,#1]{#2 ---Georgios}}
\newcommandx{\tdbNote}[2][1=]{\todo[inline,linecolor=black,backgroundcolor=red!25,bordercolor=red,#1]{#2 ---Travis}}

\newcommandx{\jeffNote}[2][1=]{\todo[inline,linecolor=black,backgroundcolor=orange!25,bordercolor=red,#1]{#2 ---Jeff}}

\newcommand{\expnumber}[2]{{#1}\mathrm{e}{#2}}

\theoremstyle{definition}
\newtheorem{definition}{Definition}[section]

\title{Contrastive Hebbian Learning with Random Feedback Weights}
\author{Georgios Detorakis, Travis Bartley, and Emre Neftci}
\date{}

\begin{document}
\maketitle

\begin{abstract}

Neural networks are commonly trained to make predictions through learning algorithms.
Contrastive Hebbian learning, which is a powerful rule inspired by gradient backpropagation, is based on Hebb's rule and the contrastive divergence 
algorithm. 
It operates in two phases, the forward (or free) phase, where the data are fed to the network, and a backward (or clamped) phase, where the target signals are 
clamped to the output layer of the network and the feedback signals are transformed through the transpose synaptic weight matrices.
This implies symmetries at the synaptic level, for which there is no evidence in the brain. 
In this work, we propose a new variant of the algorithm, called random contrastive Hebbian learning, which does not rely on any synaptic weights symmetries.
Instead, it uses random matrices to transform the feedback signals during the clamped phase, and the neural dynamics are described by first order non-linear differential equations.
The algorithm is experimentally verified by solving a Boolean logic task, classification tasks (handwritten digits and letters), and an autoencoding task.
This article also shows how the parameters affect learning, especially the random matrices. 
We use the pseudospectra analysis to investigate further how random matrices impact the learning process. 
Finally, we discuss the biological plausibility of the proposed algorithm, and how it can
give rise to better computational models for learning. 

\end{abstract}

\section{Introduction}
Learning is one of the fundamental aspects of any living organism, regardless of their
complexity. From less complex biological entities such as viruses~\cite{elde:2012},
to highly complex primates~\cite{hebb:2005,kandel:2000}, learning is of vital
importance for survival and evolution. Research in neuroscience
has dedicated significant effort to understanding the mechanisms and
principles that govern learning in highly complex organisms, such as rodents 
and primates. Both Hebbian learning~\cite{hebb:2005} and spike-timing dependent plasticity
(STDP)~\cite{bi:1998,markram:1997,zhang:1998}  
have had high impact on modern computational neuroscience, since both Hebb and STDP rules 
solve the problem of adaptation in the nervous system and can account for explaining 
synaptic plasticity~\cite{abbott:2000,cooper:2005}. 
On the other hand, the field of machine learning has made progress using gradient backpropagation (BP)~\cite{rumelhart:1988,dreyfus:1962} in
deep neural networks, providing state-of-the-art solutions to a variety of classification and representation tasks~\cite{lecun:2015}.

Despite this progress, most learning algorithms employed with artificial neural networks are implausible from a biological standpoint, this is especially true for methods based on BP. In particular,
some of the more common properties that are implausible are (i) the requirement of symmetric weights, 
(ii) neurons that do not have temporal dynamics (\emph{i.e.,} neural dynamics are not 
described by autonomous dynamical systems or maps as in the case of recurrent neural networks),
(iii) the derivatives of the non-linearities have to be computed for each layer with high
precision, (iv) the flow of information in the neural network is not similar to the one 
in real biological systems (synchronization between different phases--forward, backward--
is required), and (v) the backward phase requires the neural activity of the forward 
phase to be stored. 

In the aforementioned context there are attempts to make Machine Learning algorithms more biologically plausible. Such a biologically plausible alternative to BP is the target propagation algorithm~\cite{lee:2015}. The target propagation algorithm computes local errors at each layer of the network using information about the target, which is propagated instead of the error signal as in classical BP. However, target propagation still computes derivatives (locally) and requires symmetries at the synaptic level. Another biologically plausible learning algorithm is the recirculation algorithm~\cite{hinton:1988} and its generalization the GeneRec~\cite{oreily:1996}, where the neural network has some recurrent connections that propagate the error signals from the output layer to the hidden(s) one(s) via symmetric weights. Furthermore, the recirculation algorithm does not preserve the symmetries at the synaptic level, though GeneRec is still based on derivatives and back-propagation of error signals. Moreover, all the aforementioned algorithms require a specific pattern of information flow. Every layer should wait for the previous one to reach its equilibrium and then to proceed in receiving and processing the incoming information. This issue can be circumvented by the Contrastive Hebbian learning (CHL, deterministic Boltzmann Machine or a mean field approach)~\cite{movellan:1991,baldi:1991,xie:2003}, which is similar to the contrastive divergence algorithm~\cite{hinton:2002}. CHL is based on the Hebbian learning rule and does not require knowledge of any derivatives. Moreover, due to its non-linear continuous coupled dynamics the information flows in a continuous way. All the neural activities in all the layers may be updated simultaneously without waiting for the convergence of previous or subsequent layers. However, CHL requires synaptic symmetries since it relies on the transpose of the synaptic matrix to propagate backwards the feedback signals. 

Motivated to create a more biologically plausible CHL, we proposed random Contrastive Hebbian learning (rCHL), which avoids the use of symmetric synaptic weights, instead replacing the transpose of the synaptic weights in CHL with fixed random matrices. This was performed in a manner similar to that of Feedback Alignment (FDA)~\cite{lillicrap:2016,nokland:2016,neftci:2017}. CHL provides a good basis upon which to develop biologically realistic learning rules because it employs continuous nonlinear dynamics at the neuronal level, does not rely on gradients, allows information to flow in a coupled, synchronous way, and, is grounded upon Hebb's learning rule. CHL uses feedback to transmit information from the output layer to hidden(s) layer(s), and in instances when the feedback gain is small (such as in the clamped phase), has been demonstrated by Xie and Seung to be equivalent to BP~\cite{xie:2003}.

Using this approach, the information necessary for learning propagates backwards, though it is not transmitted through the same axons (as required in the symmetric case), but instead via separate pathways or neural populations.
Therefore, the randomness we introduce may account for the structure and dynamics of 
other cerebral areas interfering with the transmitted signals of interest or feed-back
projections as occur in the visual system~\cite{markov:2014,macknik:2009,macknik:2007,shou:2010}.
By this we do not necessarily imply that the brain implements random transformations, instead thatthe random matrices being used here in place of transpose synaptic matrices may be thought of in a more general sense as instruments for modeling unknown or hard-to-model dynamics within the brain.


The proposed learning scheme can be used in different contexts, as demonstrated on several
learning tasks. We show how rCHL can cope with (i) binary operations such as the XOR, 
(ii) classifying handwritten digits and letters, and (iii) autoencoding. In most of
these cases, the performance (in terms of the mean squared
error) of rCHL was either equivalent or similar to BP, FDA, and CHL, suggesting that rCHL
can be a potential candidate for general biologically plausible learning models. 


\section{Materials and Methods}

In this section we summarize Contrastive Hebbian learning (CHL)
\cite{xie:2003} and introduce random Contrastive Hebbian learning (rCHL). 
We assume feed-forward networks along with their corresponding feed-backs.
$L$ is the total number of layers with $1$ being the input layer and $L$ the output one.
Connections from layer $k-1$ to $k$ are given by a matrix ${\bf W}_k \in \mathbb{R}^{m\times n}$
where $m$ and $n$ are the sizes of the $k-1$ and $k$ layers (number of neurons), respectively. 
Feedback connections are given by ${\bf V}_k \in \mathbb{R}^{n \times m}$, 
which can be either a transpose matrix ${\bf W}^\intercal_k$ (in the case of CHL) or 
a random matrix ${\bf G}_k \in \mathbb{R}^{n\times m}$ in the case of random CHL.
In both CHL and rCHL, the feedback connections are multiplied by a 
constant gain, $\gamma \in \mathbb{R}$. We define the non-linearities as
Lipschitz continuous functions $f_k: \mathbb{R}^n \rightarrow \mathbb{R}^n$,
with Lipschitz constant $\alpha_k$ (\emph{i.e.} $|f_k(x) - f_k(y)| \leq \alpha_k |x - y|$,
$\forall x, y \in \mathbb{R},\, k \in \mathbb{N}$).
The state of a neuron in the $k$-th layer is described by the
state function $x^i_k: \mathbb{R} \rightarrow \mathbb{R}$, and the corresponding
bias is given by $b_k^i \in \mathbb{R}$. The dynamics of all neurons at the 
$k$-th layer are given by:
\begin{align}
	\label{eq:dyns}
	\frac{\mathrm{d}{\bf x}_k}{\mathrm{d}t} &= -{\bf x}_k +
    	f_k\big({\bf W}_k \cdot {\bf x}_{k-1} + \gamma {\bf V}_{k+1} \cdot {\bf x}_{k+1} + {\bf b}_k \big).
\end{align}

\subsection{Contrastive Hebbian Learning}

Both CHL and the proposed rCHL operate on the same principle as contrastive divergence~\cite{hinton:2002}. 
This means that learning takes place in two phases. In the positive (free) phase, the input is presented and the output is built by forward propagation
of neural activities. In the negative (clamped) phase, the outputs
are clamped, and the activity is propagated backwards towards the input layer.

In the free phase, the input layer ${\bf x}_0$ is held fixed, and the signals
are propagated forward through each layer (see the red arrows in figures~\ref{Fig:net_arch}). 
The dynamics of the neurons at each $k$-th layer are computed through the equation~\eqref{eq:dyns}
for $k=1,\ldots,L$ (the $L+1$ layer does not exist and thus
${\bf x}_{L+1} = {\bf 0}$ and ${\bf W}_{L+1} = {\bf 0}$). During the clamped phase, the target
signal is clamped at the output
layer ${\bf x}_L$ and the activity of all the neurons in every layer is computed through
equation~\eqref{eq:dyns} for $k=1,\ldots,L-1$ (notice here that the input layer $k=0$ does
not express any dynamics). The backward flow is illustrated as cyan arrows in figure~\ref{Fig:net_arch}.
At the end of the two phases we update the synaptic weights and the
biases based on the following equations, 
\begin{subequations}
\label{eq:total_w}
\begin{align}
    \Delta {\bf W}_k &= \eta \gamma^{k-L} \big( \hat{{\bf x}}_k \otimes \hat{{\bf x}}_{k-1} - \check{{\bf x}}_k \otimes \check{{\bf x}}_{k-1} \big)   , \quad k=1,\ldots,L, \\
    \Delta {\bf b}_k &= \eta \gamma^{k-L} \Big( \hat{{\bf x}}_k - \check{{\bf x}}_k  \Big)
\end{align}
\end{subequations}
where $\otimes$ is the tensor product, $\eta$ is the learning rate, $\gamma$ 
is the feedback gain, and $\Delta {\bf W}_k$ is weight update. $\check{{\bf x}}_k$
represents the activity of neurons in the $k$-th layer at the equilibrium 
configuration of equation~\eqref{eq:dyns} in the free phase and
$\hat{{\bf x}}_k$ the activity of the $k$-th layer in the clamped phase.  

\subsection{Random Contrastive Hebbian Learning}

As described by equation~\eqref{eq:dyns}, CHL implicitly requires
the synaptic weights to be symmetric (${\bf W}_k^\intercal$) in order to 
use the feedback information. In this work, the main contribution is to 
cast aside the symmetry and replace all the transpose matrices that appear
in CHL with random matrices ${\bf G}_k$. This idea is similar 
to random feedback alignment~\cite{lillicrap:2016,nokland:2016},
where the error signals are propagated back through random matrices
that remain constant during learning. Therefore,
equation~\eqref{eq:dyns} is modified to the following:
\begin{align}
	\label{eq:ran_dyns}
	\frac{\mathrm{d}{\bf x}_k}{\mathrm{d}t} &= -{\bf x}_k +
    	f_k\big({\bf W}_k \cdot {\bf x}_{k-1} + \gamma {\bf G}_{k+1} \cdot {\bf x}_{k+1} + {\bf b}_k \big),
\end{align}
where the learning increments remain the same. In order to properly 
apply CHL and rCHL, we follow the second strategy of training 
proposed by Movellan in \cite{movellan:1991} (pg. $12$, case $2$), 
suggesting to first let activity settle during the clamped phase. Then, without
resetting activations, free the output units and allow activity to settle again. This 
method assures that when the minimum for the clamped phase has been 
reached, it remains stable.

We summarize rCHL in Algorithm~\ref{algo:rchl},
where $\mathcal{I}$ is the input dataset, $\mathcal{T}$ is the corresponding
target set, $N$ is the number of input samples, and $L$ is the number
of layers.
\begin{algorithm}[!h]
	\begin{algorithmic}
    	\Require $\mathcal{I}$, $\mathcal{T}$, $L$, epochs, $t_f$, $dt$
        \Ensure ${\bf W}$
        \State $N = |\mathcal{I}|$
        \State Initialize ${\bf G}_k$ and ${\bf W}_k$ randomly for $k=1,\ldots,L$
        \If {\text{Bias is adaptive}}
        	\State Initialize ${\bf b}_k$ randomly
        \Else
        	\State ${\bf b}_k$ is fixed
        \EndIf
        \For{$e \gets 1, \ldots, \text{epochs}$}
        	\State $i \sim \mathcal{U}(1, N)$
        	\State $\check{{\bf x}}_0 = \mathcal{I}_i$
            \State $\hat{{\bf x}}_L = \mathcal{T}_i$
            \For{$k \gets 1, \ldots, L-1$}
               	\For{$n \gets 1, \ldots, t_f$ with step $dt$} \Comment{Forward Phase}
                	\State $\hat{{\bf x}}_k[n] = \hat{{\bf x}}_k[n-1] + dt \Big(-\hat{{\bf x}}_k[n-1] +
            		f_k\big({\bf W}_k \cdot \hat{{\bf x}}_{k-1}[n-1] +
                    \gamma {\bf G}_{k+1} \cdot \hat{{\bf x}}_{k+1}[n-1] +
                	{\bf b}_k \big) \Big)$
                \EndFor
            \EndFor
            \For{$k \gets 1, \ldots, L$}
            	\For{$n \gets 1, \ldots, t_f$ with step $dt$} \Comment{Backward Phase}
                	\State $\check{{\bf x}}_k[n] = \check{{\bf x}}_k[n-1] + dt \Big(
                    -\check{{\bf x}}_k[n-1] + f_k\big({\bf W}_k \cdot \check{{\bf x}}_{k-1}[n-1]
                    + \gamma {\bf G}_{k+1} \cdot \check{{\bf x}}_{k+1}[n-1] + {\bf b}_k \big) \Big)$
                \EndFor
            \EndFor
            \For{$k \gets 1, \ldots, L$}
            	\State ${\bf W}_k \leftarrow {\bf W}_k + \eta \gamma^{k-L} \big( \hat{{\bf x}}_k
            	\hat{{\bf x}}_{k-1}^\intercal - \check{{\bf x}}_k \check{{\bf x}}_{k-1}^\intercal \big)$
                \If {\text{Bias is adaptive}}
                	\State ${\bf b}_k \leftarrow {\bf b}_k + \eta \gamma^{k-L}
                    	\big( \hat{{\bf x}}_k - \check{{\bf x}}_k  \big)$
                \EndIf
            \EndFor
        \EndFor
	\end{algorithmic}
\caption{Random contrastive Hebbian learning (rCHL). $\mathcal{I}$ is the input
dataset, $\mathcal{T}$ is the corresponding target set (labels) of the input data
set, and $L$ is the number of layers of the network, $t_f$ is the simulation time 
and $dt$ the Forward Euler method's time-step.}
\label{algo:rchl}
\end{algorithm}
The rCHL starts with randomly initializing the synaptic weights and the feed-back
random matrices. If the bias is not allowed to learn then it is fixed at the 
very beginning. If it's permitted to adapt then it is randomly initialized. 
Then in every epoch rCHL picks up randomly an input sample and assigns
it to the input layer, $\check{{\bf x}}_0$. At the same time, it assigns the corresponding 
label (target) to the output layer $\hat{{\bf x}}_L$. Then it solves all the non-linear 
coupled differential equations for each layer using a Forward Euler method for the 
backward phase. This means that it computes the $\hat{{\bf x}}_k$ and then it solves
again the system of the coupled non-linear equations for the forward phase in order to
compute the $\check{{\bf x}}_k$. Once all the activities for the forward and the backward
phases have been computed, rCHL updates the synaptic weights and the biases,
if they are allowed to be updated, based on equations~\eqref{eq:total_w}a and b.

\begin{figure}[htpb!]
	\centering
    \includegraphics[width=0.5\textwidth]{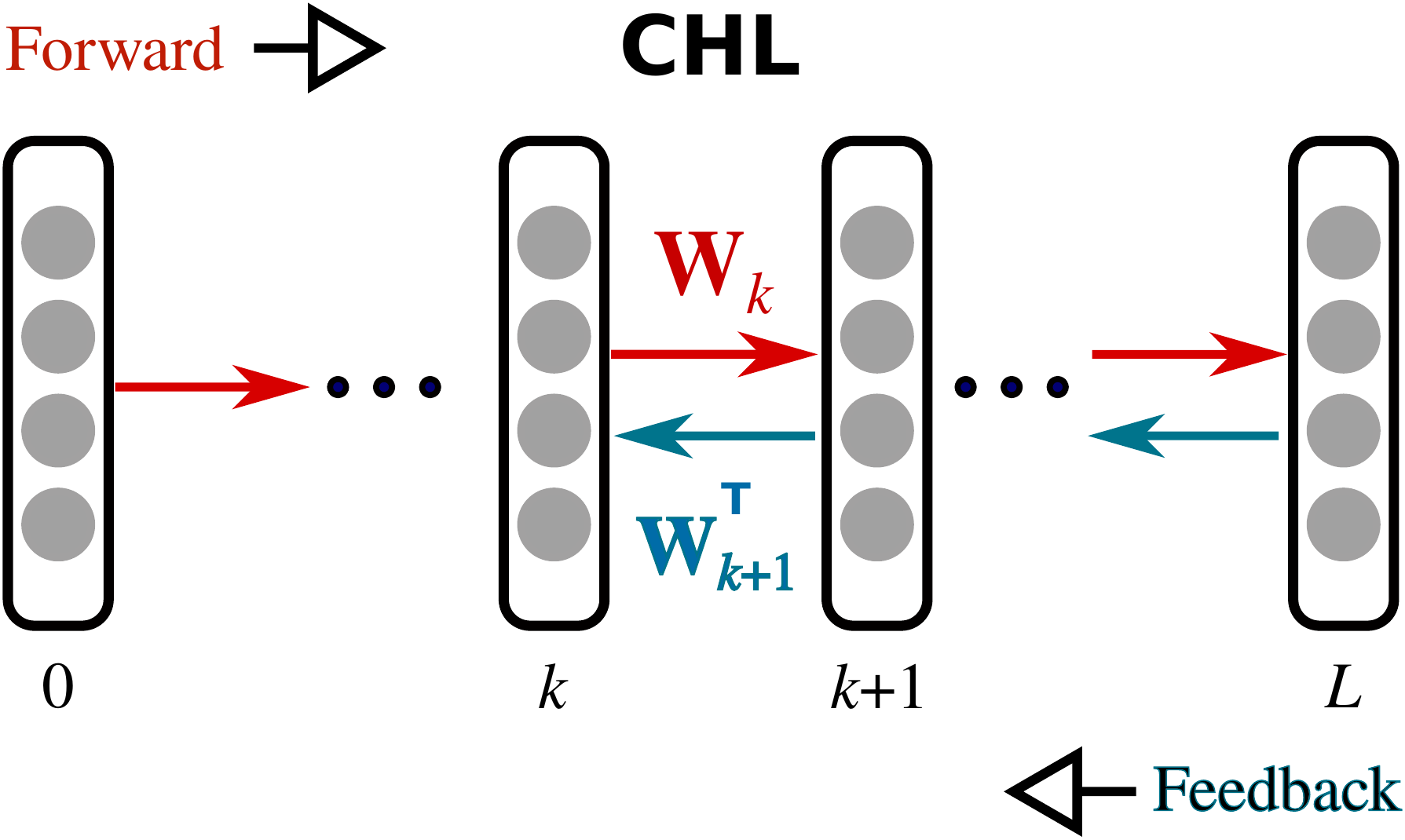}
    \includegraphics[width=0.5\textwidth]{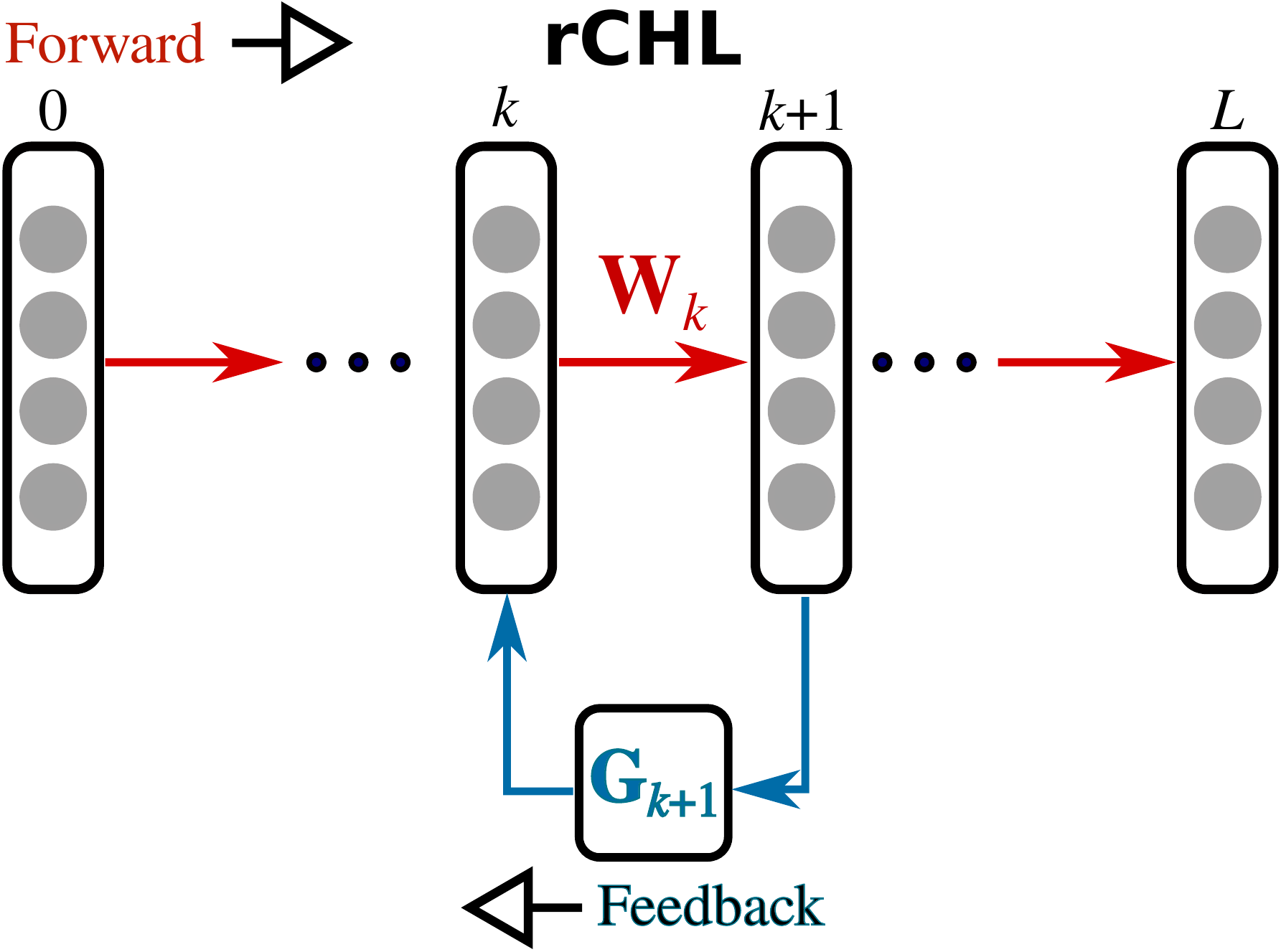}
    \caption{{\bfseries \sffamily Learning scheme information flow.} CHL (top panel) and 
    rCHL (lower panel) are illustrated in this figure. Both CHL and rCHL consist of two phases.
    In the forward phase, the input signal is fed to the network, and the activity propagates up
    to the deepest layer through matrices ${\bf W}_k$ (red arrows). In the backward phase, the
    output (target) signal is clamped, whilst the input signal is still present and affects 
    the neural dynamics. The activity is propagated backwards through matrix 
    ${\bf W}^{\intercal}_{k+1}$ (navy color arrows) in the CHL and ${\bf G}_{k+1}$ in the rCHL.
    It is clear that rCHL's feedback mechanisms does not require any symmetries and acts more like
    a feedback system on the dynamics of neurons conveying information from the $k+1$-th
    layer back to $k$-th layer.}
    \label{Fig:net_arch}
\end{figure}

Figure~\ref{Fig:net_arch} illustrates the neural network architecture and the information
flow of CHL (top panel) and rCHL (bottom panel). In the forward pass, the input is provided
and information is propagated through matrices ${\bf W}_k$. In the backward phase, the output 
is clamped, and the information flows from the output layer to the hidden(s) through the 
transpose matrix ${\bf W}_{k+1}^\intercal$ (CHL) or random matrices ${\bf G}_{k+1}$ (rCHL).

\section{Results}
Next, we demonstrate rCHL on a variety of tasks. The algorithm successfully solves
logical operations, classification tasks and pattern generation problems. 
In the logical operations and classification tasks we compared our results against the
state-of-the-art back-propagation (BP) and feed-back alignment (FDA). 
In all rCHL and CHL simulations, we used the following settings, unless otherwise stated: 
time step $dt=0.08$, total simulation time $t_f = 30\mathrm{ms}$, learning rate $\eta=0.1$,
and feedback gain $\gamma = 0.05$.

\subsection{Bars and Stripes Classification}

\begin{figure}[htpb!]
\centering
	\includegraphics[width=0.9\textwidth]{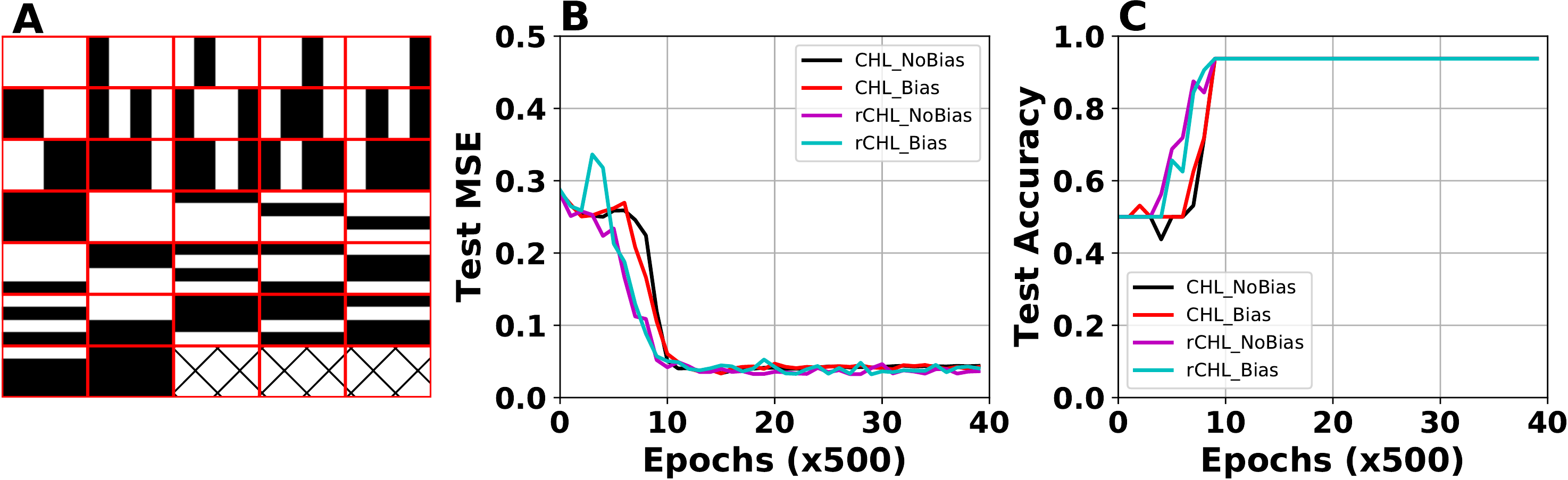}
    \caption{{\bfseries \sffamily Bars and stripes classification with rCHL}.
    {\bfseries \sffamily A} Bars and stripes dataset used to train the neural network.
    {\bfseries \sffamily B} the test MSE computed every $500$ epochs. In every epoch, 
    a stimulus (image) is presented to the network. Here, the rCHL algorithm (cyan
    and magenta curves) are compared against the CHL algorithm (black and red curves). 
    After $5,000$ epochs the network has converged.  
    {\bfseries \sffamily C} Test accuracy of the network on classifying the bars
    and stripes. We presented the entire dataset ($32$ images) during testing.}
\label{Fig:2}
\end{figure}

We investigate how the parameters of the rCHL affect the learning 
process. In particular, we examine how to select the random matrix $\bf G$, the feedback
gain $\gamma$, and the number of layers $L$. To this end we use the bars and stripes
classification task to demonstrate the effect of the different parameters~\cite{mackay:2003}.
The dataset consists of $32$ binary images (black--$0$ and white--$1$) of size $4\times 4$
representing bars and stripes, as shown in figure~\ref{Fig:2}{\bfseries \sffamily A}. We
train a network of of three layers ($L=3$) with sizes $16-50-2$ to classify the input data
into bars and stripes. During each epoch we pick up randomly one out of $32$ images and 
present it to the network for $20,000$ epochs. Every $500$ epochs we freeze the learning 
and we test the performance of the network. We measure the Mean Square Error (MSE)
(\emph{i.e.}, $\frac{1}{N} \sum_{i=1}^{N}(y_i - \hat{y}_i)^2$, where ${\bf y}$ is a 
reference signal of $N$ components and $\hat{{\bf y}}$ is the estimated signal)
and the accuracy of the network.   
All neurons have sigmoid activation functions ($f_k(x) = \frac{1}{1+\exp(-x)}$
for all $k=1, \ldots, L$). 

The achieved test MSE is shown in figure~\ref{Fig:2}{\bfseries \sffamily B}
and the test accuracy in figure~\ref{Fig:2}{\bfseries \sffamily C}. It is apparent 
that rCHL converges, and the binary classification task has been learned after $5,000$ epochs.
We tested two different versions of rCHL, first we used the bias terms
(cyan color in figure~\ref{Fig:2}) in equation~\eqref{eq:ran_dyns}. All ${\bf b}_k$ 
have been initialized randomly from a uniform distribution $\mathcal{U}(-0.1, 0.1)$
and they are allowed to learn based on equation~\eqref{eq:total_w}. 
In the second case we set the bias terms to zero (purple color in figure \ref{Fig:2}).
The same method was also followed for CHL.
Figures \ref{Fig:2}{\bfseries \sffamily B} and \ref{Fig:2}{\bfseries \sffamily C} indicate 
that CHL and rCHL can achieve similar results in the task. 
Once we have established the functionality of the rCHL, we investigate how the 
parameters of the random matrix $\bf G$, the feedback gain $\gamma$, the 
learning rate $\eta$, and the number of layers $L$ affect the learning.
Therefore, we can conclude at that point that the rCHL has similar behavior with 
the CHL and the learning process works as well as CHL's. In the following paragraphs
we are investigating how four basic parameters of the model (learning rate, feed-back
gain, number of layers, and the random matrix) affect the learning using the Bars 
and Stripes classification task as toy model. We choose to not use the bias terms 
since they do not affect the learning in this particular task. 

\subsubsection{Feedback Gain} 

\begin{figure}[htpb!]
\centering
	\includegraphics[width=0.9\textwidth]{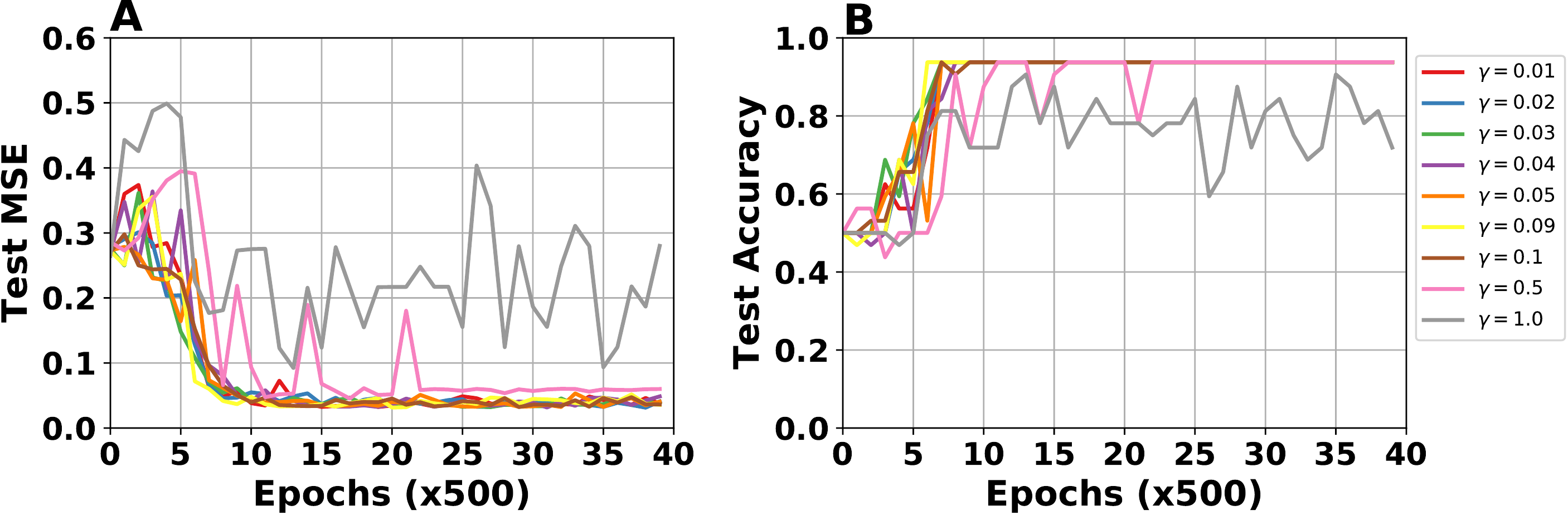}
    \caption{{\bfseries \sffamily Effect of feedback gain $\gamma$ on bars and stripes classification}.
    This figure illustrates the test MSE {\bfseries \sffamily A} and the test accuracy 
    {\bfseries \sffamily B} for nine different values of feedback gain $\gamma$. Both panels illustrate that the neural network trained using the rCHL algorithm
    learns to classify the bars and stripes dataset in all cases, except where $\gamma=1.0$.
    In this case, the MSE is high and the classification is unstable.}
\label{Fig:3}
\end{figure}

First we start with the feedback gain by
sweeping it over the interval $[0.01, 1.0]$, and drawing the initial values
of the synaptic weights as well as the random matrix from a uniform 
distribution $\mathcal{U}(-0.5, 0.5)$. The effect of the feedback gain for 
standard CHL has been examined 
in \cite{xie:2003}. In figures~\ref{Fig:3}{\bfseries \sffamily A} and {\bfseries \sffamily B},
the test MSE and accuracy are shown for different values of 
$\gamma$. Lower feedback gain (about $0.1$), works well for
rCHL. Even when the feedback gain is around $0.0001$ the learning process 
still works (data not shown) and the convergence is fast. This is explained
by the fact that the term $\gamma^{k-L}$ in equation~\eqref{eq:total_w}(a) 
becomes extremely large for small feedback gains for the very first layers and 
quite small for the deeper layers. On the other hand, when when the 
feedback gain is high (\emph{e.g.}, $\gamma=1.0$) the rCHL does not 
converge (gray color in figure~\ref{Fig:3}). 

\subsubsection{Learning Rate}

\begin{figure}[htpb!]
\centering
	\includegraphics[width=0.9\textwidth]{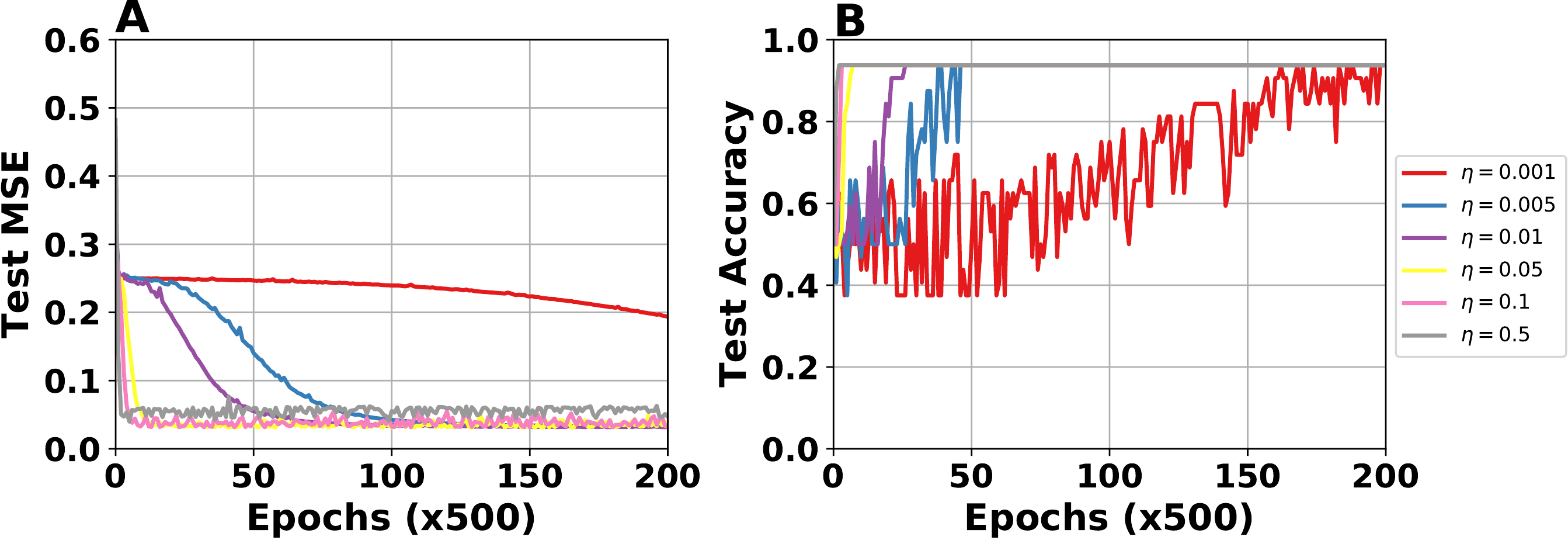}
    \caption{{\bfseries \sffamily Effect of learning rate $\eta$ on bars and stripes classification}.
     This figure illustrates the test MSE {\bfseries \sffamily A} and the test accuracy 
     {\bfseries \sffamily B} for six different values of learning rate $\eta$. It is 
     apparent that if the learning rate is too low the learning process takes more time
     to converge. In this case we keep the feed-back gain constant at $\gamma=0.5$. }
\label{Fig:19}
\end{figure}

Another important parameter that affects learning is the rate the rCHL adjusts the 
synaptic weights, or the learning rate $\eta$. Therefore, we use the same network 
architecture $16-50-2$ and we keep fix the feedback gain $\gamma=0.05$ and we vary 
the learning rate, \emph{i.e.} $\eta = \{0.001, 0.005, 0.01, 0.05, 0.1, 0.5 \}$. 
Figure~\ref{Fig:19} shows the test error and accuracy for the various values of 
learning rate. When the learning rate is too low the learning diverges. On the
other hand when the learning rate is higher the learning converges faster. For 
the values between $0.005$ and $0.01$ the learning converges smoother but takes 
more time to reach the equilibrium in comparison to the higher values. In this case
the random matrix and the synaptic weights have been initialized by a uniform 
distribution $\mathcal{U}(-0.5, 0.5)$. 

\subsubsection{Number of Layers} 

\begin{figure}[htpb!]
\centering
	\includegraphics[width=0.9\textwidth]{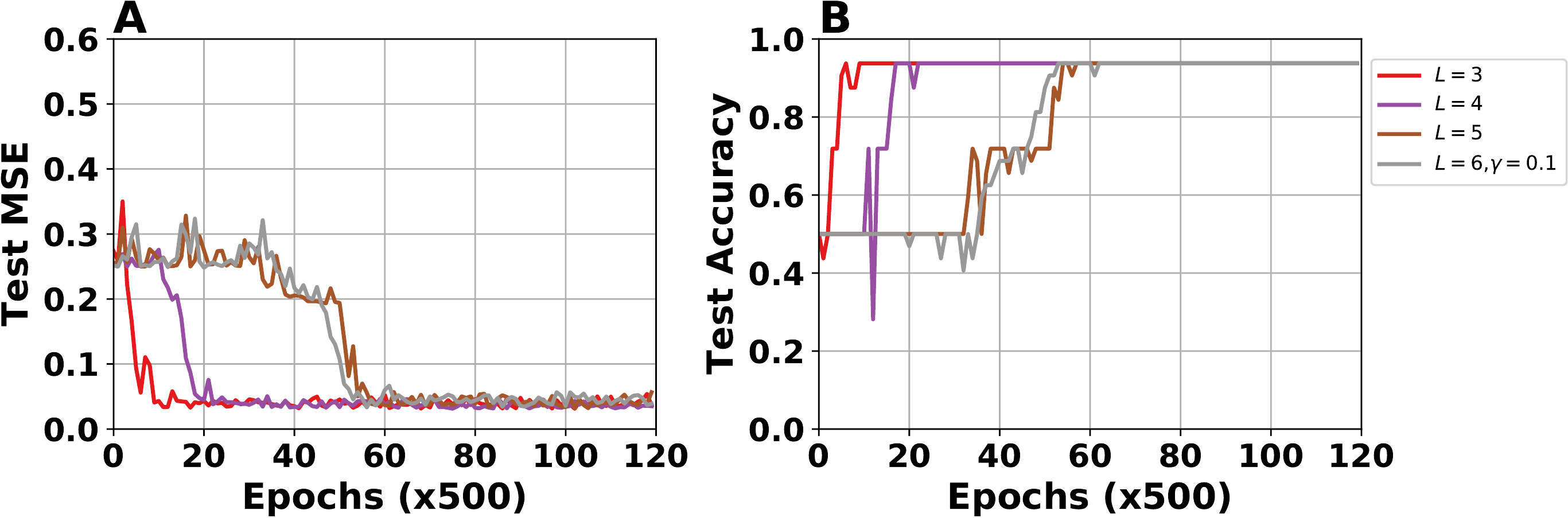}
    \caption{{\bfseries \sffamily Effect of number of layers $L$ on bars and stripes classification}.
    This figure illustrates the test MSE {\bfseries \sffamily A} and the test accuracy 
    {\bfseries \sffamily B} for four different values of $L$, which represents the number of layers. For the largest value $L=6$, rCHL diverges (data not shown) and we have 
    to re-tune the feedback gain. Once the feedback gain has been increased, the convergence
    of rCHL is guaranteed (gray line).}
\label{Fig:4}
\end{figure}

Next, we investigate how the number of hidden layers
affects learning. To this end, we train four different neural networks using rCHL. 
The configurations for the networks are $16-50-2$ ($L=3$),
$16-50-10-2$ ($L=4$), $16-50-20-10-2$ ($L=5$), and $16-50-30-20-10-2$ ($L=6$). 
As before, we draw the synaptic weights and the random matrix from a uniform distribution
$\mathcal{U}(-0.5, 0.5)$. 
Figures~\ref{Fig:4} {\bfseries \sffamily A} and {\bfseries \sffamily B} show 
the test MSE and test accuracy of the networks. As we increase the 
number of layers, the rCHL networks fail to converge (data not shown). However, when 
we increase the feedback gain $\gamma$, convergence is achievable 
(gray curves in figure~\ref{Fig:4}).
This behavior can be explained by the
fact that the feedback gain affects the synaptic weight update by a factor 
$\gamma ^ {k-L}$. This means that the more layers a network has,
the higher $\gamma$ should be.

\subsubsection{Feedback Random Matrix}

The random feedback matrix plays a crucial role in the rCHL learning process. 
It conveys the information from the output layer back to the hidden(s) one(s). 
Therefore, the feedback random matrix, essentially, alters the signals by 
applying the feedback from layer $k+1$ to neural dynamics at layer $k$. 
Therefore, the learning process is directly affected by the choice of the
feedback random matrices within the network. The properties of random matrices
arise from the distributions that generate them. In this case we investigate
how random matrices generated by a normal $\mathcal{N}(0, \sigma)$ and a 
uniform $\mathcal{U}(a, b)$ distribution can impact the learning process. 
Furthermore, we define the length of the uniform distribution interval as
$\ell = |b| + |a|$. 

\begin{figure}[htpb!]
\centering
	\includegraphics[width=0.95\textwidth]{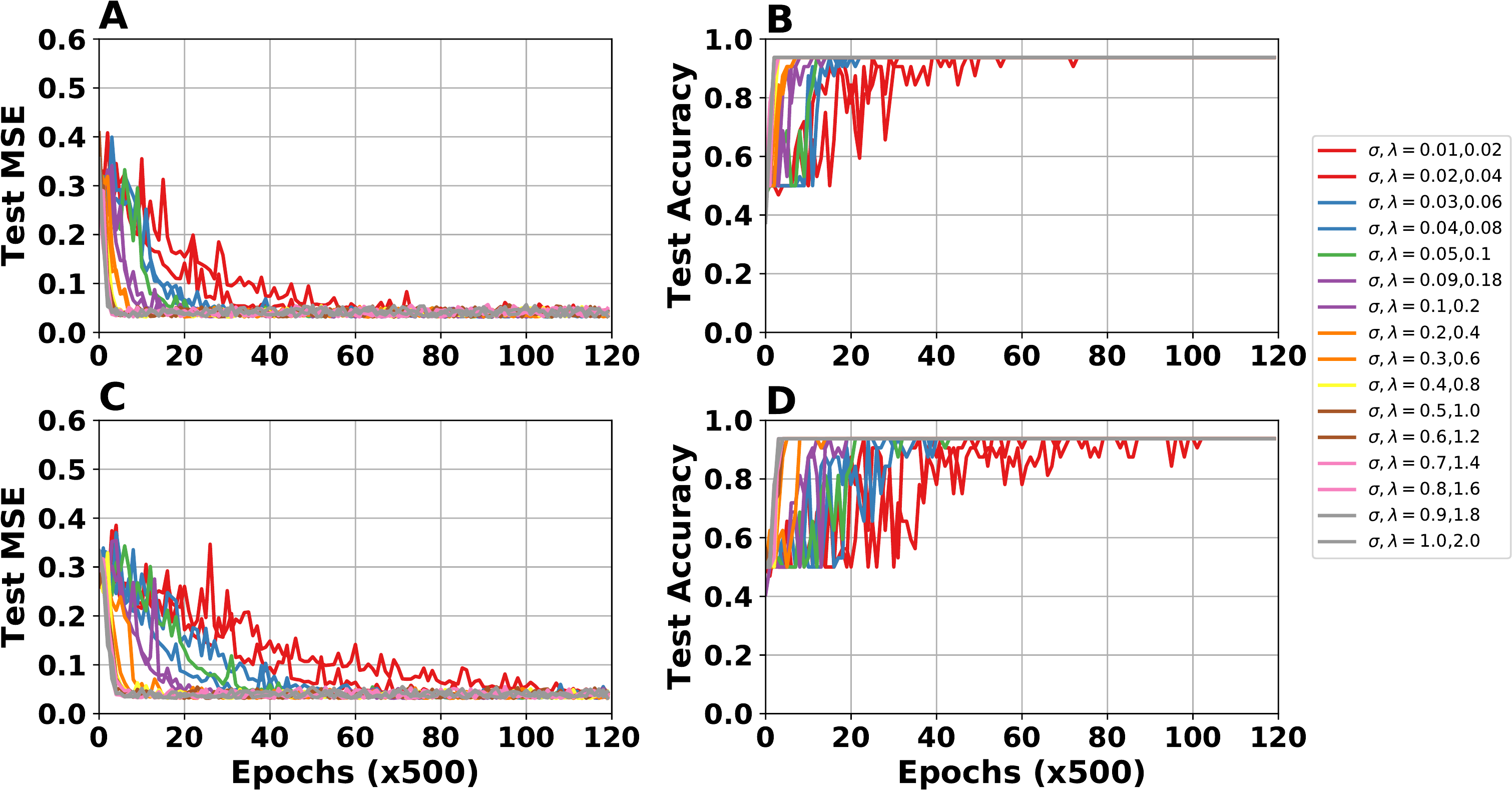}
    \caption{{\bfseries \sffamily Effect of matrix distribution on bars and stripes classification.} This figure
    illustrates how the random matrix ${\bf G}_2$ affects the learning process when it
    is initialized from a normal and a uniform distribution. 
    {\bfseries \sffamily A} Test MSE for the normal distribution, 
    {\bfseries \sffamily B} corresponding test accuracy. 
    {\bfseries \sffamily C} Test MSE for the uniform distribution,
    {\bfseries \sffamily D} corresponding test accuracy.
    The colored lines indicate different values for the variance $\sigma$ of the normal
    distribution and the interval (length $\ell$) of the uniform distribution (see the
    legend). The higher the 
    variance or the length, the faster and more stable the convergence of the learning.
    The network with configuration $16-50-2$ performs a classification task on the bars and stripes dataset
    (see the text for more details).}
    \label{fig:distros}
\end{figure}

We start by varying the variance $\sigma$, and the length $\ell$ of the
two different distributions. Next the network is evaluated on the bars and 
stripes classification task, and the MSE and accuracy are recorded. 
Figures~\ref{fig:distros}{\bfseries \sffamily A} and {\bfseries \sffamily B} 
show the test MSE and accuracy, for sixteen different 
$\sigma$ values for the normal distribution. As shown, the higher 
the variance, the faster convergence is reached. Figures~\ref{fig:distros}
{\bfseries \sffamily C} and {\bfseries \sffamily D} indicate the test MSE and 
accuracy for sixteen different values of $\ell$ of the uniform distribution. 
In this case, we observe that the shorter the interval (smaller $\ell$),
the slower the convergence.
Meanwhile, the wider the interval, the faster and  the convergence is.
The convergence for short intervals is slower for the uniform distribution than for
corresponding small variance values for the normal distribution 
(see figures~\ref{fig:distros}{\bfseries \sffamily A} and {\bfseries \sffamily C}). 

One of the key aspects in Random Matrix Theory is the spectrum (\emph{i.e.,} eigenvalues)
of the random matrix and especially the distribution of the eigenvalues~\cite{vu:2014}.
However, most connection matrices in neural networks are not square 
and non-normal (\emph{i.e.}, ${\bf V}{\bf V}^{*} \neq {\bf V}^{*}{\bf V}$, 
where ${\bf V} \in \mathbb{R}^{n \times n}$). Therefore, one alternative to study the
spectrum of random matrices that are rectangular and non-normal is to study their
singular values and the corresponding $\lambda_{\epsilon}$-pseudospectra
~\cite{trefethen:1991,trefethen:2005,wright:2002} (we provide 
a brief description of the $\lambda_{\epsilon}$-pseudospectra in the 
Appendix~\ref{appendix:pseudo}). The pseudospectra define a set of 
pseudoeigenvalues ($\lambda$) of a matrix ${\bf A} \in \mathbb{R}^{m \times n}$,
if for some eigenvector $\pmb{\upsilon}$ with $||\pmb{\upsilon}|| = 1$ it holds
$||({\bf A} - \lambda\tilde{{\bf I}})\pmb{\upsilon}|| \leq \epsilon$. Hence
pseudospectra indicates potential sensitivity of eigenvalues under perturbations 
of the matrix ${\bf A}$. In this study we are interested in identifying spectral 
properties that can be related to the learning (\emph{i.e.,} convergence,
speed of convergence, oscillations). 

\begin{figure}[htpb!]
\centering
	\includegraphics[width=0.8\textwidth]{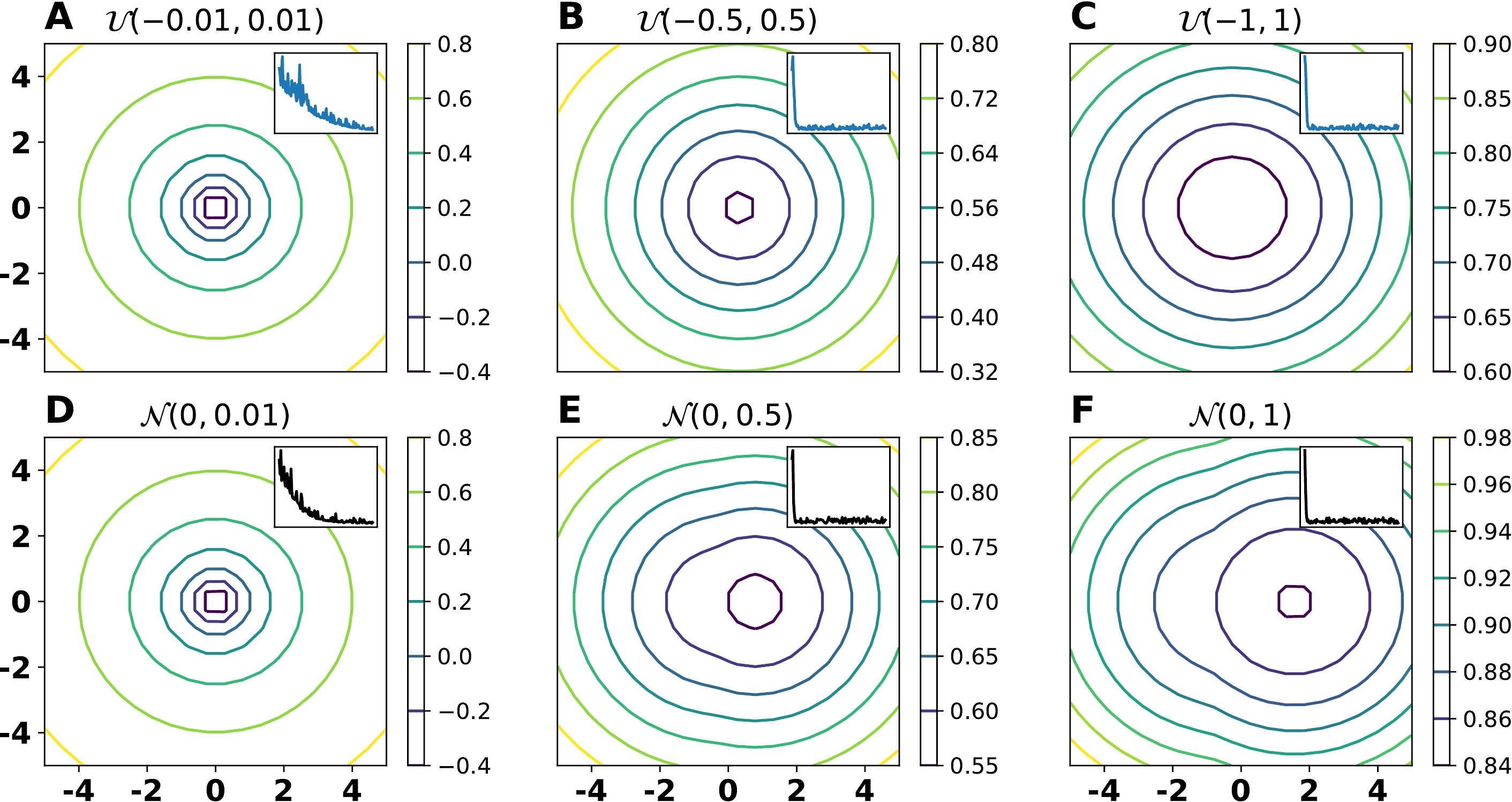}
    \caption{{\bfseries \sffamily $\Lambda_{\epsilon}$-pseudospectra of ${\bf G}_2$ for the Bars and Stripes classification.} Six different feedback random matrices have been
    analyzed using the $\Lambda_{\epsilon}$-pseudospectra method. The colormap indicates
    the different $\epsilon$ values. The top row shows the $\Lambda_{\epsilon}$ for 
    the matrices drawn by uniform distributions from intervals
    ({\bfseries \sffamily A}) $(-0.01, 0.01)$,
    ({\bfseries \sffamily B}) $(-0.5, 0.5)$, and
    ({\bfseries \sffamily C}) $(-1, 1)$. 
    The bottom row illustrates $\Lambda_{\epsilon}$ for the random matrices drawn 
    from normal distributions with zero mean and variances
    ({\bfseries \sffamily D}) $\sigma=0.01$,
    ({\bfseries \sffamily E}) $\sigma=0.5$, 
    and ({\bfseries \sffamily F}) $\sigma=1$. }
    \label{fig:psd}
\end{figure}

Figure~\ref{fig:psd} illustrates the $\lambda_{\epsilon}$-pseudospectra for the 
random matrix ${\bf G}_2$ for the Bars and Stripes classification task. We chose 
three cases for the uniform and three for the normal distribution from figure~\ref{fig:distros}
and we then applied on the random matrices the algorithm given in~\cite{wright:2002} to 
compute the $\lambda_{\epsilon}$-pseudospectra of ${\bf G}_2$ in each case. In every 
panel the contour lines indicate the minimum singular values that correspond to different 
values of $\log_{10}(\epsilon)$ on the complex plane. In addition in every panel we 
provide the corresponding test MSE (inset plot). Subplots~\ref{fig:psd}{\bfseries \sffamily A}
{\bfseries \sffamily B} and {\bfseries \sffamily C} depict the pseudospectra of ${\bf G}_2$ drawn
from uniform distributions ($\mathcal{U}(-0.01, 0.01)$, $\mathcal{U}(-0.5, 0.5)$, $\mathcal{U}(-1, 1)$, 
respectively), as well as the test MSE. Likewise, subplots {\bfseries \sffamily D}, 
{\bfseries \sffamily E}, and {\bfseries \sffamily F} illustrate the cases of normal 
distributions ($\mathcal{N}(0, 0.01)$, $\mathcal{N}(0, 0.5)$, $\mathcal{N}(0, 1)$, 
respectively).

Since the only varying parameter in this experiment is the way we generate
the random matrix the learning process is solely affected by that matrix. Therefore, 
in the cases {\bfseries \sffamily A} and {\bfseries \sffamily D} the pseudospectra 
of the two different distributions look identical, whilst the test MSE has similar 
behavior, it decays slower (blue and black lines in the insets). In both cases the 
minimum value for $\epsilon$ is the same (and around the origin). In the other cases
{\bfseries \sffamily B}, {\bfseries \sffamily C}, {\bfseries \sffamily E}, and 
{\bfseries \sffamily F} the convergence of the test MSE toward zero is faster (inset)
and less violent (in terms of oscillations). The pseudospectra show higher minimum 
values for $\epsilon$. For the uniform distribution all the $\epsilon$ values are 
arranged on concentric circles. On the other hand,  the normal distribution with 
variances $\sigma=0.5$ and $\sigma=1$ causes a shift towards the right-half complex
plane of the pseudospectrum. The minimum $\epsilon$ is higher in comparison to 
the uniform cases. One more remark is that for these four cases the uniform 
distributions have smaller $\epsilon$ values in comparison to their normal 
counterparts. This might lead to a less oscillatory behavior of the test MSE
as it is shown in the insets (blue and black curves). 
In all cases, we compute the pseudospectra using the implementation provided in~\cite{wright:2002}\footnote{
The source code can be found at: \url{https://github.com/gdetor/pygpsa}}.

\subsection{Exclusive Or (XOR)}

The exclusive or (denoted XOR or $\oplus$) problem consists of the evaluation of four 
possible Boolean input states (\emph{i.e.}, $1\oplus0=1$, $1\oplus1=0$, 
$0\oplus1=1$, $0\oplus 0=0$). 
To solve the XOR problem, we use a feed-forward network with one hidden layer
and one output layer ($L=3$) with a configuration $2-2-1$. All neural units are sigmoidal: 
$f(x) = (1+\exp(-x))^{-1}$, and we train the neural network on the XOR problem
for $5,000$ epochs using CHL and rCHL. 
The random matrix $\bf G$ for rCHL has been initialized from a uniform 
distribution $\mathcal{U}(-0.2, 0.2)$. In every epoch we present $128$ samples
to the network and every $500$ epochs we measure the MSE and the accuracy on the
test dataset. The accuracy here is defined to be
$\frac{1}{4}\sum_{i=1}^{4} |x_2(t) - T_i| < \epsilon$, 
where $\epsilon = 0.01$ and the index $i$ runs over the test samples. 
\begin{figure}[!htpb]
	\centering
    \includegraphics[width=0.8\textwidth]{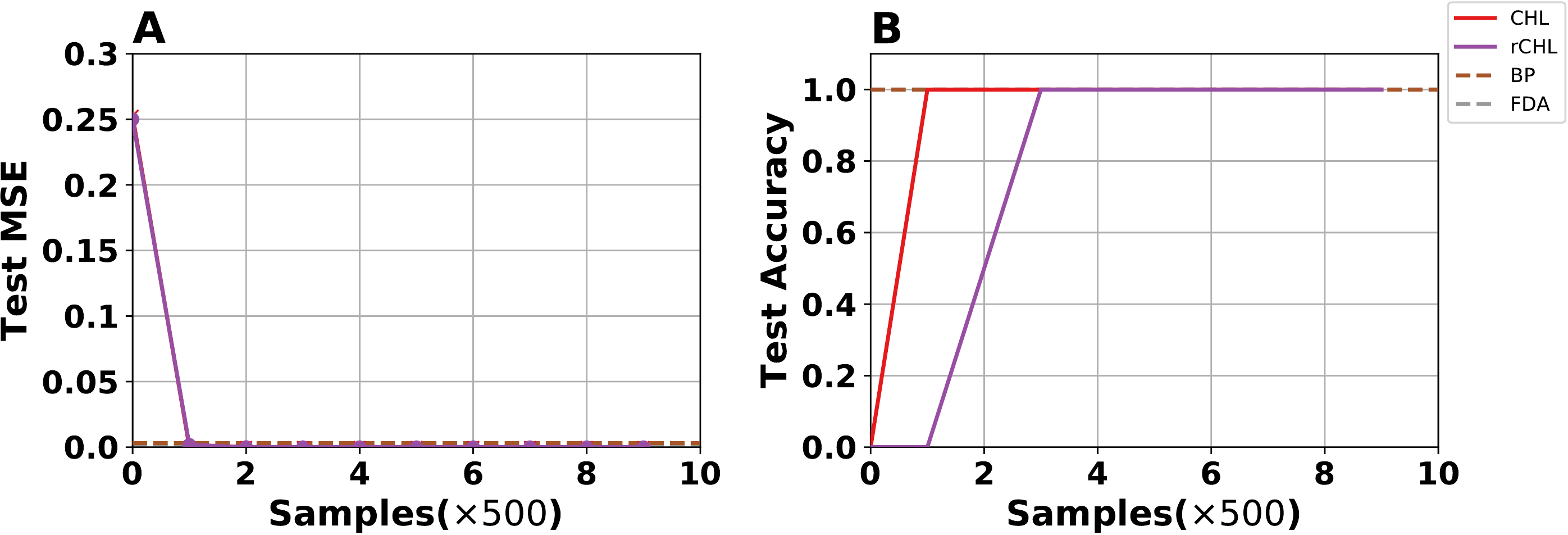}
    \caption{{\bfseries \sffamily Exclusive OR (XOR).} Four different learning algorithms 
    have been use to train a feed-forward network with three layers, $L=3$, ($2-2-1$).
    ({\bfseries \sffamily A}) Test MSE error illustrated against the number of samples,
    ({\bfseries \sffamily B}) test accuracy. The red and purples lines indicate the CHL
    and rCHL, respectively. The brown and gray dashed lines show the minimum test MSE 
    and the maximum test accuracy for the BP and the FDA, respectively.}
    \label{Fig:xor_error}
\end{figure}

\begin{figure}[htpb!]
	\centering
    \includegraphics[width=0.8\textwidth]{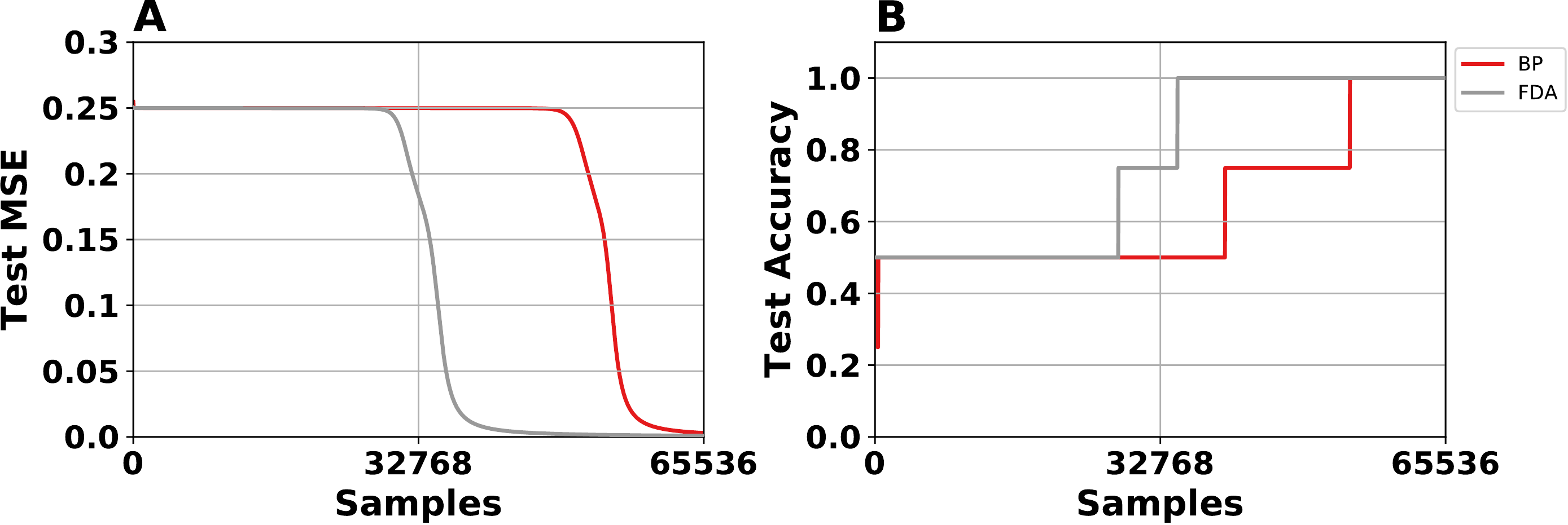}
    \caption{{\bfseries \sffamily Backpropagation (BP) and feedback alignment (FDA) on XOR.}
    The test MSE ({\bfseries \sffamily A}) and the test accuracy ({\bfseries \sffamily B}) show
    the convergence of BP (red) and FDA (gray) in solving the XOR problem. For training both
    BP and FDA, we used $65,536$ batches, and each batch had a size of $1$ sample. The 
    learning rate was $\eta = 0.1$ and the momentum $\alpha = 0.1$. Synaptic weights and 
    biases were initialized using uniform distributions $\mathcal{U}(-0.3, 0.3)$ and 
    $\mathcal{U}(-0.1, 0.1)$, respectively.}
    \label{Fig:si_xor}
\end{figure}

The results of the learning are shown in figure~\ref{Fig:xor_error},
where figure~\ref{Fig:xor_error}{\bfseries \sffamily A} shows the test error,
and figure~\ref{Fig:xor_error}{\sffamily \bfseries B} the test accuracy 
for CHL (red line) and rCHL (purple line). The test error
and the test accuracy attained is the same for BP (brown dashed line) and FDA 
(gray dashed line). Comparing against the error and accuracy of BP and FDA
(see SI figure~\ref{Fig:si_xor}), rCHL converges faster and more smoothly. This is because rCHL and CHL are on-line learning algorithms, and the
input and target signals are both embedded into the dynamics of the neurons. This 
leads to a rapid convergence of the Hebbian learning rule, which rapidly assimilates the proper associations between input and output signals.

\subsection{Handwritten Digit and Letter Classification}

\begin{figure}[htpb!]
	\centering
    \includegraphics[width=0.8\textwidth]{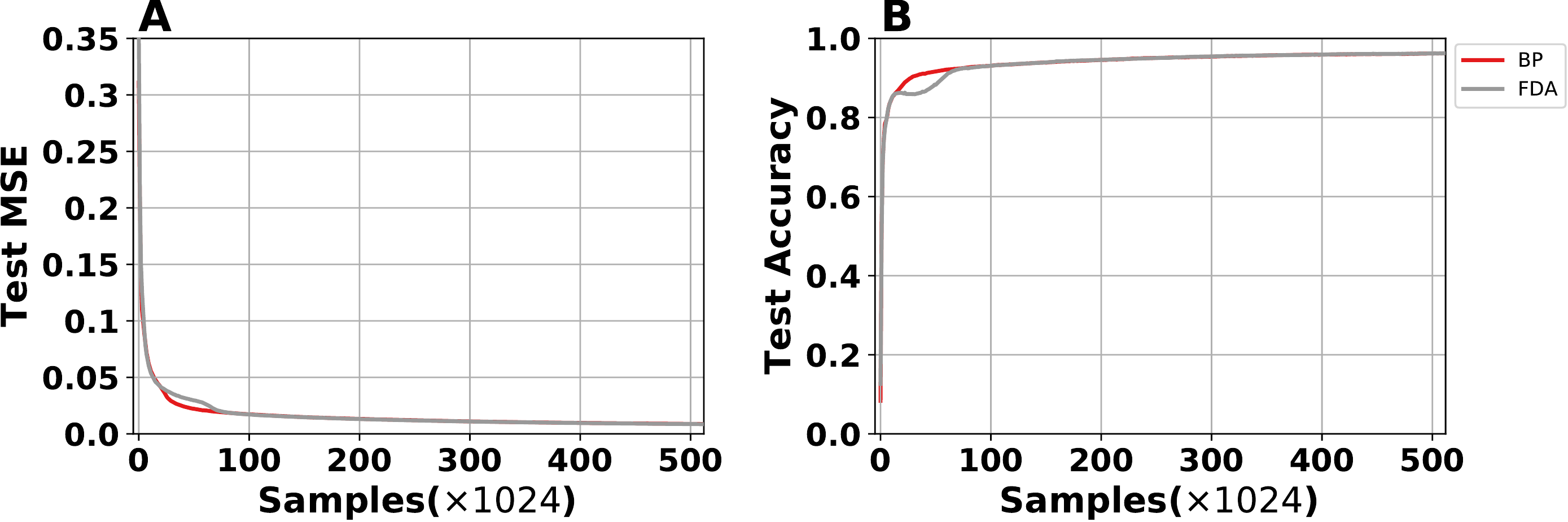}
    \caption{{\bfseries \sffamily Backpropagation (BP) and feedback alignment (FDA) on MNIST.}
    The test MSE ({\bfseries \sffamily A}) and the test accuracy ({\bfseries \sffamily B}) show
    the convergence of BP (red) and FDA (gray) in solving MNIST handwritten digits 
    classification problem. For training both BP and FDA, we used $524,288$ batches, and
    each batch had a size of $128$ samples. The 
    learning rate was $\eta = 0.07$. Synaptic weights and biases were initialized using 
    uniform distributions $\mathcal{U}(-0.3, 0.3)$ and $\mathcal{U}(-0.1, 0.1)$, respectively.}
    \label{Fig:si_mnist}
\end{figure}

For the classification tasks, we used the MNIST~\cite{deng:2012} 
dataset of $10$ handwritten digits and the eMNIST datasets~\cite{cohen:2017}
of $26$ handwritten letters of the English alphabet. 
The neural network layouts for the MNIST and the eMNIST was $784-128-64-10$ and  
$784-256-128-26$, respectively. On every unit in both MNIST and eMNIST 
networks, we use a sigmoid function: $f(x) = (1+\exp(-x))^{-1}$. 
We drew the initial synaptic weights and the random matrices from uniform distributions $\mathcal{U}(-0.6, 0.6)$ and 
$\mathcal{U}(-0.3, 0.3)$, respectively. 
We trained the network for $20$ (MNIST) and $40$ (eMNIST) epochs, and in each epoch we
present the entire MNIST and eMNIST datasets, which consist of $60,000$ and $124,800$ 
images, respectively. At the end of each epoch, we measured the MSE and 
the accuracy of the network. The accuracy is defined to be the ratio
of the successfully classified images to the total presented test images ($10,000$
for MNIST and $20,800$ for eMNIST, respectively). 
\begin{figure}[!htpb]
	\centering
    \includegraphics[width=0.8\textwidth]{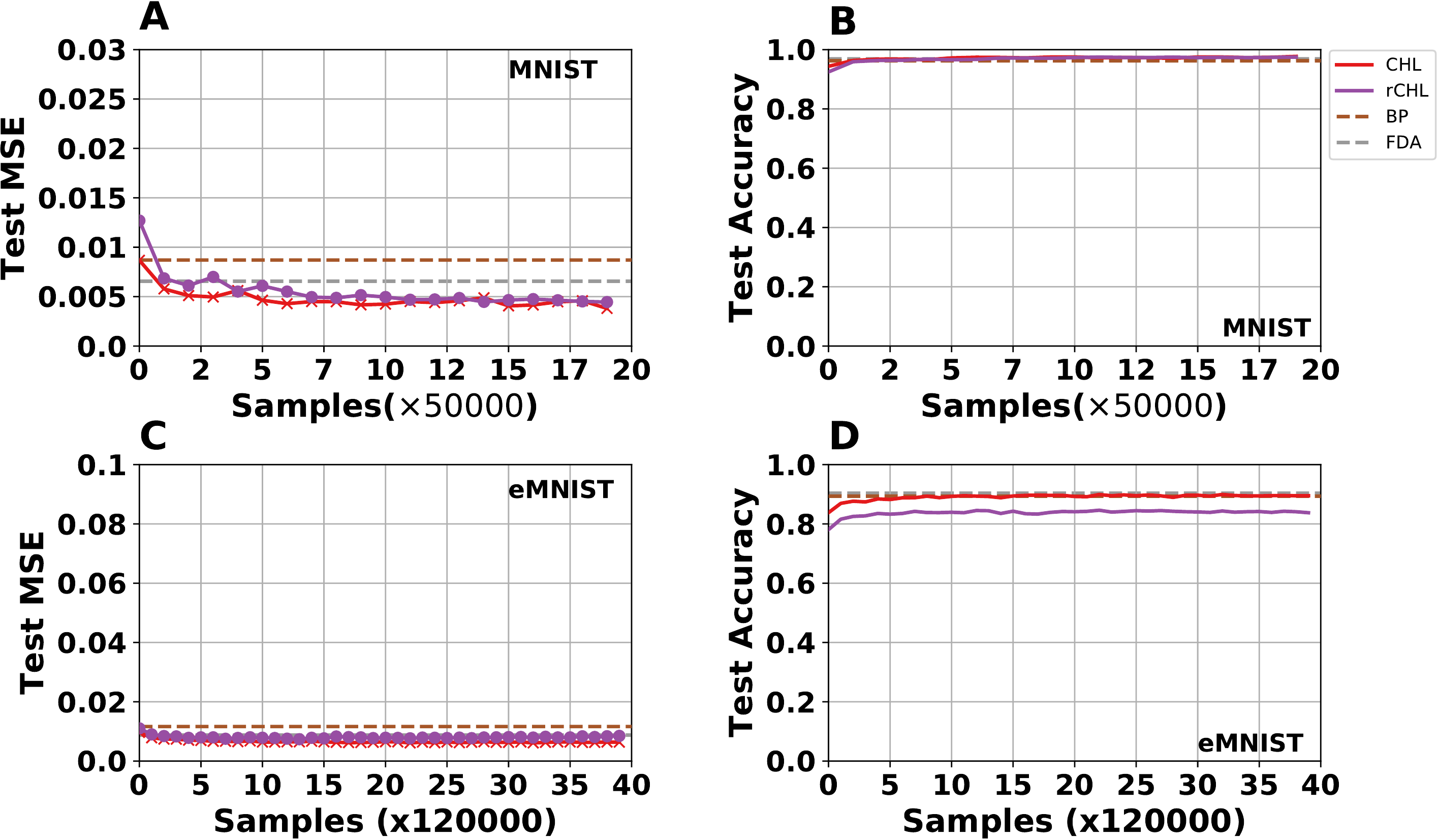}
    \caption{{\bfseries \sffamily MNIST and eMNIST digits classification.}
    The figure illustrates \
    ({\bfseries \sffamily A}) the test error and
    ({\bfseries \sffamily B}) the accuracy of a neural network ($784-128-64-10$)
    with sigmoid function as non-linearity in all layers.
    The network was trained on the entire MNIST set of digits ($50,000$ images)
    and tested on the whole MNIST test set ($10,000$ images). 
    In addition, ({\bfseries \sffamily C}) illustrates the test error and
    ({\bfseries \sffamily D}) the accuracy of a neural network ($784-256-128-26$)
    with sigmoid function as non-linearity in all layers. The network was trained on the 
    entire eMNIST set of digits ($124,800$ images) and tested on the whole eMNIST
    test set ($20,800$ images). In both cases the MNIST and the eMNIST, CHL and rCHL
    have similar performance (red and purple curves, respectively). }
    \label{Fig:mnist_err}
\end{figure}
Figure~\ref{Fig:mnist_err}{\bfseries \sffamily A} illustrates the test error 
of CHL (red line) and rCHL (purple line), respectively. The error 
is lower than BP and FDA (brown and gray dashed lines, respectively),
and the convergence is faster (compared against SI figure~\ref{Fig:si_mnist}, 
where BP and FDA test MSE and accuracy are illustrated). However, the classification
test accuracy is the same as for BP and FDA, as figure~\ref{Fig:mnist_err}{\bfseries \sffamily B}
indicates. The convergence of the BP and FDA algorithms can be seen in SI 
figure~\ref{Fig:si_mnist}, as well as the test accuracy of those algorithms
performing on the same type of neural network (same neural units and network
architecture).

\begin{figure}[htpb!]
	\centering
    \includegraphics[width=0.8\textwidth]{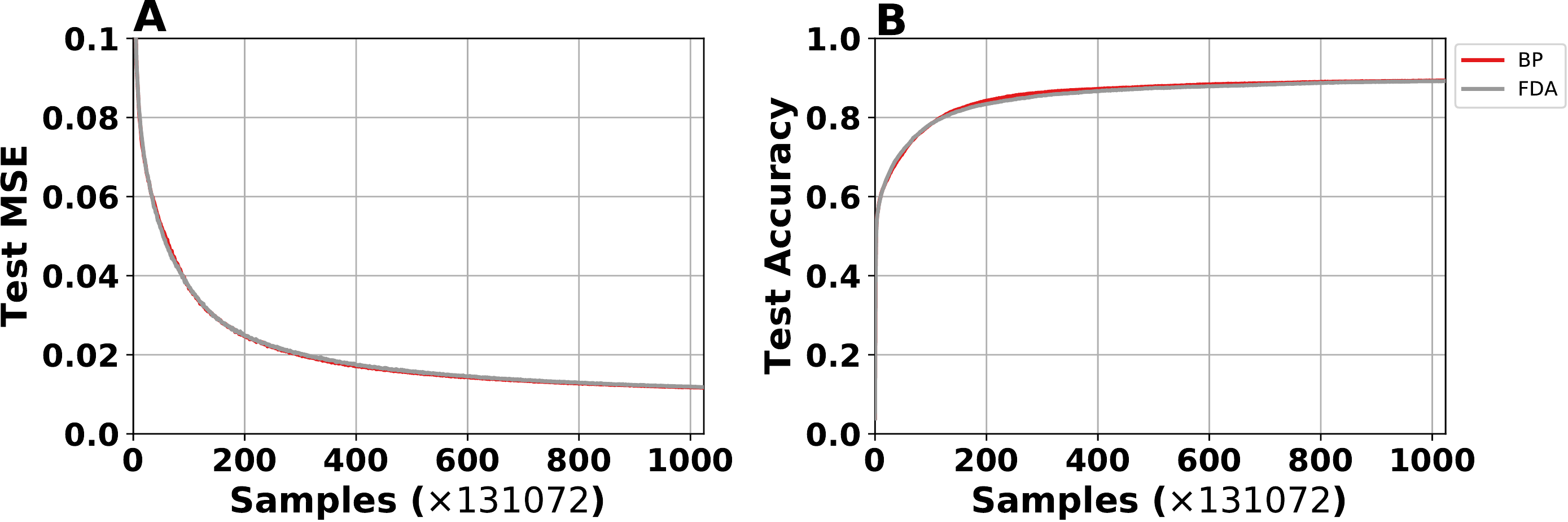}
    \caption{{\bfseries \sffamily Backpropagation (BP) and feedback alignment (FDA) on eMNIST.}
    The test MSE ({\bfseries \sffamily A}) and the test accuracy ({\bfseries \sffamily B}) show
    the convergence of BP (red) and FDA (gray) in solving the eMNIST handwritten letters 
    classification problem ($26$ letters of the English alphabet). For training both BP and
    FDA, we used $1,048,576$ batches, and each batch had size of $128$ samples. The 
    learning rate was $\eta = 0.07$. Synaptic weights and biases were initialized using 
    uniform distributions $\mathcal{U}(-0.3, 0.3)$ and $\mathcal{U}(-0.1, 0.1)$, respectively.}
    \label{Fig:si_emnist}
\end{figure}

For the eMNIST dataset, the test error is illustrated in 
figure~\ref{Fig:mnist_err}{\bfseries \sffamily C}, where the error of
CHL (red line) and rCHL (purple line) are close to the error attained by BP and FDA (brown and gray dashed lines). However, the classification test accuracy of rCHL is worse than the other three
learning algorithms (CHL, BP and FDA), as figures~\ref{Fig:mnist_err}{\bfseries \sffamily D} and
SI \ref{Fig:si_emnist} {\bfseries \sffamily B} show. For the eMNIST data set the accuracy 
of the rCHL is lower than the other three algorithms despite the fact that the error is smaller 
than the errors of BP and FDA. This drawback might be due to the feed-back random matrix and 
the lack of symmetric synaptic connections. This is also supported by the fact that the CHL 
(which differs from rCHL only in the way the feedback signals are transmitted) achieves an 
accuracy as good as the BP and FDA (minimum test error and maximum test accuracy for all four
algorithms are provided in table~\ref{table:0},Appendix~\ref{appendix:table_err_acc}).

\subsection{Autoencoder}
The final test case was the implementation of an autoencoder using CHL and
rCHL. An autoencoder is a neural network that learns to reconstruct the input 
data using a restricted latent representation. 
Classic autoencoders consist of three layers: the first layer 
is called the encoder, which is responsible for encoding the input data to the 
hidden layer, which describes a code. The third layer is the decoder, which 
generates the approximation of the input data, based on the code provided by the
hidden layer \cite{goodfellow:2016}. 

To this end, we use a network with three layers $L=3$, with a 
layout $784-36-784$, and we use the MNIST handwritten digits. This means that
we encode the input images of dimension $784$ to a dimension of $36$. 
For both CHL and rCHL, the learning rate is set to $\eta=0.05$. The synaptic weights
and the random matrix ${\bf G}$ were initialized from uniform distributions 
$\mathcal{U}(-0.5, 0.5)$ and $\mathcal{U}(-1, 1)$, respectively. 
We trained the network for $20$ epochs, and in each epoch we presented to the network
$5,000$ samples of MNIST handwritten digits.
\begin{figure}[htpb!]
\centering
	\includegraphics[width=0.5\textwidth]{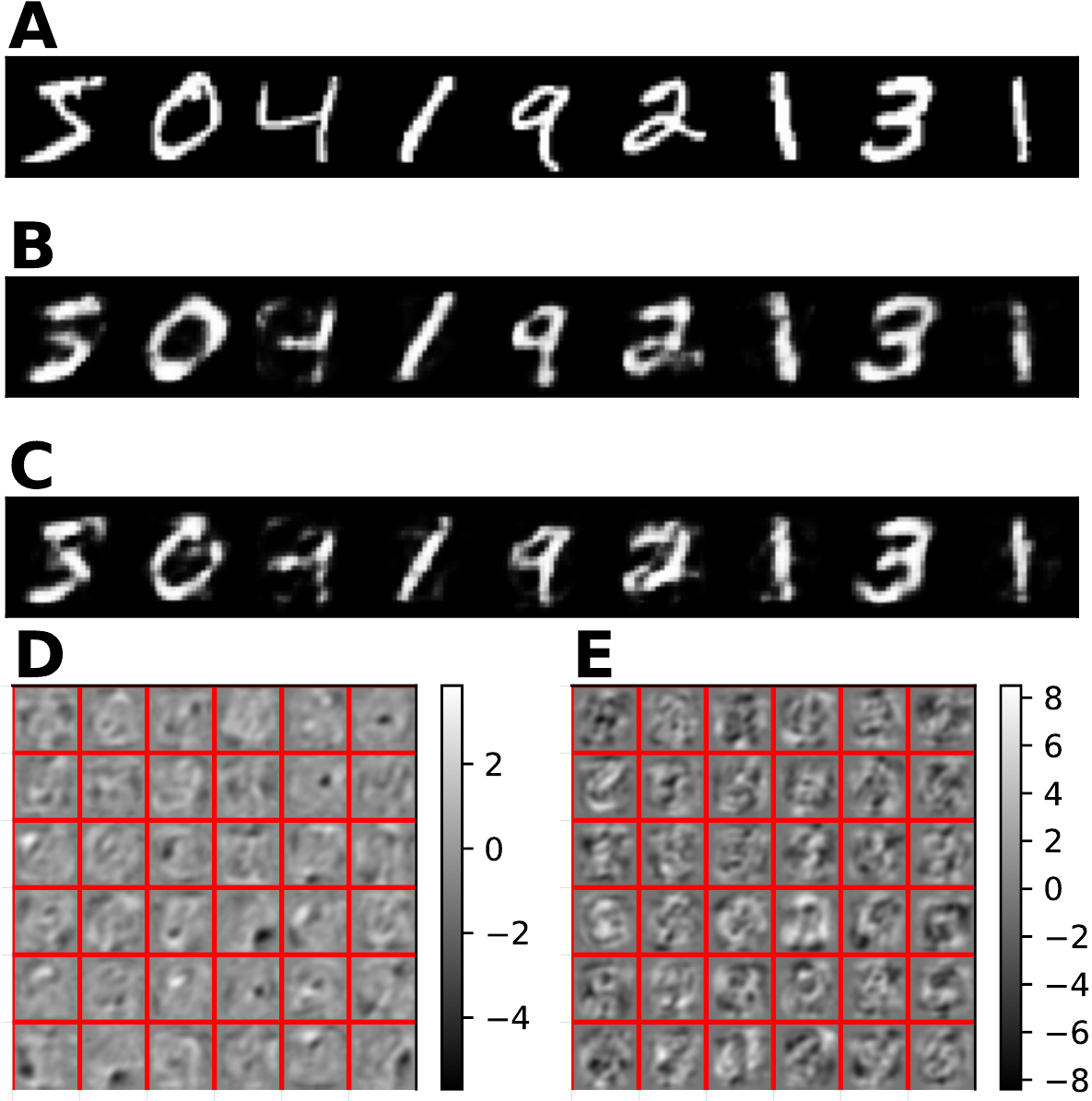}
    \caption{{\bfseries \sffamily CHL and rCHL autoencoder.}
    The MNIST dataset was used to train an autoencoder using CHL
    and rCHL. ({\bfseries \sffamily A}) shows some randomly chosen MNIST
    digits. After training the network for $40$ epochs (during each epoch
    we presented $5,000$ MNIST images to the network), it was able to reconstruct
    the learned digits as panel ({\bfseries \sffamily B}) shows for CHL,
    and panel ({\bfseries \sffamily C}) for rCHL. The learned representations
    are shown in panels ({\bfseries \sffamily D}) and ({\bfseries \sffamily E})
    for CHL and rCHL, respectively.}
\label{Fig:15}
\end{figure}

After training the autoencoder, we fed $9$ digits 
(figure~\ref{Fig:15}{\bfseries \sffamily A}) from the MNIST dataset to the autoencoders. The decoding (reconstruction) of the input
data is illustrated in figures~\ref{Fig:15}{\bfseries \sffamily B} and 
\ref{Fig:15}{\bfseries \sffamily C} for CHL and rCHL,
respectively. The results indicate that rCHL and
its non-random counterpart (CHL), can both learn to approximate the identity
function, and thus they are both good candidates for implementing autoencoders.
The reconstruction (decoding) was not perfect, which implies that the network 
correctly learned an approximation (and not the exact function), indicating 
that the network had learned the appropriate representations (code). The 
representations (codebooks) are shown in figures~\ref{Fig:15}{\bfseries \sffamily D}
and \ref{Fig:15}{\bfseries \sffamily E} for the CHL and rCHL models, respectively. 
As expected, the representations of rCHL were more noisy than that of the 
CHL model. This can be partially explained by the chosen simulation parameters in
this experiment.

\section{Discussion}
In this work, we introduced a modified version of Contrastive Hebbian Learning 
~\cite{xie:2003,movellan:1991,baldi:1991}, based on random feedback connections.
The key contribution of this work is a biologically plausible variant of CHL, attained by 
using random fixed matrices
(during learning) instead of bidirectional synapses. 
We have shown that this new variation of CHL, random Contrastive Hebbian
Learning, can achieve results equivalent to that of CHL, BP, 
or FDA. In addition, rCHL can solve
logical problems such as the XOR, as well as classification tasks, such as the
handwritten digits and letters classification (MNIST and eMNIST datasets).
Furthermore, rCHL supports representation learning (\emph{e.g.}, autoencoders),
as well. 

We provide a thorough investigation of the algorithm, examining how
the different parameters of the learning scheme affect the learning process. 
We have shown that the feedback gain $\gamma$ affects the performance of learning.
The smaller the gain $\gamma$, the better the performance of the algorithm, 
implying that the feedback gain should be small. This is in accordance with
previous theoretical results found in~\cite{xie:2003}, where CHL requires
low values of $\gamma$ in order to be equivalent to BP. A second factor that
affects learning is the number of hidden layers. When the number of hidden
layers increases, the performance of the learning 
decreases. However, this can be mitigated by increasing the feedback gain
$\gamma$ (see figure~\ref{Fig:4}).
Finally, we tested the initial conditions of the random matrix ${\bf G}$. 
We found that if we choose the random matrix from a normal distribution with 
zero mean, the convergence speed of the learning algorithm is faster than if 
we draw ${\bf G}$ from a uniform distribution. In addition, the variance
$\sigma$ and the interval length $\lambda$ of the normal and uniform 
distributions, respectively, affect the learning as well. The smaller the values,
the faster the convergence of the learning. 

We further investigated the feedback random matrices using tools provided
by the pseudospectra analysis~\cite{trefethen:1991,wright:2002}. The convergence
of learning can be affected by the choice of distribution that generates the 
feedback random matrix. We found that sub-Gaussian feedback matrices
with normal-like pseudospectrum tend to cause slower convergence of learning,
but a more robust one. On the other hand, Gaussian random matrices tend to have faster 
convergence. 

Both CHL and rCHL share some common properties such as (i) the neurons express
non-linear dynamics, (ii) the learning rule is based on Hebb's rule~\cite{hebb:2005}, 
which means no biologically implausible information is necessary (\emph{e.g.},
knowledge of the derivatives of the non-linearities of the neural units),
(iii) the propagation of the activity from one layer to the next takes place in 
a manner similar that of biological systems. All the layers are coupled through
a non-linear dynamical system, which implies that when an input is presented the algorithm does not have to wait until all the units in the $k$-th 
layer have fired. Instead, neurons can propagate their activity through couplings
to the next $(k+1)$-th layer. This is not the case for artificial neural networks
where the activity is computed for all the units of the $k$-th layer prior to transmission to the subsequent layer. Thus, CHL and rCHL circulate signals in 
a more natural fashion (similar to the processes evident in biological nervous systems). 

The salient difference between CHL and rCHL is the replacement of symmetric 
synaptic weights (CHL) with fixed random matrices (rCHL). Therefore, the feedback
connections do not require any symmetric weights (transpose matrices) of the
feed-forward connections. This is in accordance with biology since there is no evidence thus far to indicate that chemical synapses (one of the two major synapse
types in the nervous system, the other being electrical synapses)
are bidirectional~\cite{kandel:2000,pickel:2013}. 
More precisely, we can interpret the random feedback as a model of afferent 
feedback connections from higher hierarchical layers to lower ones. 
For instance, primary sensory areas are connected in series with higher sensory
or multi-modal areas. These higher areas project back to the primary areas through
feedback pathways~\cite{markov:2012,oh:2014,glasser:2016}. Other types 
of feedback pathways to which our approach may be considered relevant are the inter-laminar connections within 
cortical layers~\cite{harris:2013,thomson:2003}. 
Therefore, such feedback pathways can be modeled as random feedback matrices 
that interfere with the neural dynamics of the lower layers (top-down signal transmission).
The randomness in these feedback pathways can account for currently unknown underlying 
neural dynamics or random networks with interesting properties~\cite{may:1976,vu:2014}.

Our current implementation, rCHL, allows us to interpret the backward 
phase (clamped) as a top-down signal propagation. For instance, when a 
visual stimulus is presented, such as a letter, and a subject is required to learn the 
semantic of the letter (which letter is presented), then the target can be 
the semantic and the input signal the image of the letter. Therefore, we
can assume that the forward phase simulates the bottom-up signal propagation
from the primary sensory cortices to higher associate cortices, and the 
backward phase as the top-down signal propagation from the higher cognitive
areas back to the lower cortical areas. 

Due to the nature of the learning algorithm and the two phases (positive and negative),
the input and the target signals are clamped, and the dynamics of those signals
are captured and embedded in the neural dynamics of the network. This can
be compared to target propagation~\cite{lee:2015,le:1986}, 
where the loss gradient of BP is replaced by a target value. When the 
target value is very close to the neural activation of the forward pass, 
TP behaves like BP\@. These sort of learning algorithms solve the credit 
assignment problem~\cite{minsky:1961}, using local information. In the 
proposed model, the target and the input signals are both embedded into the
dynamics of the neural activity and affect the learning process in an indirect
way through a Hebb and an anti-Hebb rule.

The major contribution of this work is the development of a new method to 
implement CHL using random feedback instead of symmetric one. This leads 
to a more biologically plausible implementation of CHL without loss of performance 
and accuracy in most of the cases studied. In addition, the algorithm offers a shorter
runtime since it does not require the computation of the transpose matrix for synaptic 
weights at every time step. Furthermore, the proposed algorithm 
offers a suitable learning scheme for neuromorphic devices since its neural 
dynamics can be transformed into spiking neurons~\cite{gerstner:2002}. Therefore,
we can build spiking neural networks with STDP-like learning rules as equivalent
to the Hebb's rule in rCHL and use ideas from event-based Contrastive 
Divergence~\cite{neftci:2014} and synaptic sampling machines~\cite{neftci:2016}
in order to implement a neuromorphic rCHL. This idea is similar to the 
event-based random back-propagation algorithm~\cite{neftci:2017}, where 
the authors have implemented an event-based feedback alignment equivalent 
for neuromorphic devices.

A potential extension of the model might be a replacement of the firing rate 
units with spiking neurons, such as leaky integrate-and-fire units, in order to 
make the computations even more similar to the biological case. Another potential
direction would be to remove entirely the synchronization in the neural dynamics 
integration. This means that the integrations of neural dynamics can take place 
on-demand in a more event-based fashion~\cite{rougier:2011,taouali:2009}. 
Since the units are coupled with each other, and the propagation of activity
takes place in a more natural way, we might be able to use asynchronous 
algorithms for implementing rCHL such that it scales up in a more efficient
and natural way. 
Another extension of the model in the future would be to impose sparsity
constraints on the learning rule in order to render data encoding processes 
more efficient. Furthermore, such a modification would make the model able to 
simulate more biological phenomena, such as the sparse compression occurring within the hippocampus~\cite{petrantonakis:2014}. 

In the future, we would like to conduct more analytical work to determine the optimal
type of random matrix ${\bf G}$, for a given problem, and how to design such a matrix,
if it does exist. This means that if an optimal 
random matrix ${\bf G}$ exists, then one can properly choose the eigenvalues
(or singular values) and the distribution that generates that matrix based 
on the problem they would like to solve. To this end, the pseudospectra analysis 
can play a key role. Pseudospectra can help us to better understand the relations
between singular values, convergence of learning and accuracy. This can lead to
the development of sophisticated methods for designing feedback matrices with 
particular properties. Therefore, we could improve and accelerate the learning 
process.

\subsection*{Author Contributions}
GD conceived the idea, implemented CHL and rCHL, ran the CHL and
rCHL experiments; TB implemented and ran the BP and FDA experiments;
GD analyzed and interpreted the results; All the authors wrote and 
reviewed the manuscript.

\subsection*{Acknowledgments}
This work was supported in part by the Intel Corporation and by the National
Science Foundation under grant 1640081. We thank NVIDIA corporation for providing the GPU
used in this work.

\section{Appendix}

\subsection{$\Lambda_{\epsilon}$-pseudospectra}
\label{appendix:pseudo}

Let ${\bf V} \in \mathbb{R}^{m \times n}$ be a rectangular matrix, not necessarily 
normal (\emph{i.e.}, ${\bf V}{\bf V}^{*} \neq {\bf V}^{*}{\bf V}$). The 
$\epsilon$-pseudospectra ($\Lambda_{\epsilon}$)~\cite{trefethen:2005,wright:2002}
is the set of $\epsilon$-eigenvalues a closed subset of $\mathbb{C}$. 
\begin{definition}[$\epsilon$-Pseudospectra]
	\label{def:psd1}
    Let ${\bf V} \in \mathbb{C} \text{ or } \mathbb{R}$, 
    $z \in \mathbb{C}$ and $\tilde{{\bf I}} \in \mathbb{R}^{m \times n}$ (the identity 
    matrix with ones on the main diagonal and zeros elsewhere), then the $\epsilon$-pseudospectra
    $\Lambda_{\epsilon} = \{z \in \mathbb{C}: ||({\bf V} - z\tilde{{\bf I}}){\pmb \upsilon}||\leq \epsilon
    \text{ for some } \pmb{\upsilon} \in \mathbb{C}^n, ||\pmb{\upsilon}|| = 1\}$.
\end{definition}
\begin{definition}[$\epsilon$-Pseudospectra]
	\label{def:psd2}
    Let ${\bf V} \in \mathbb{C} \text{ or } \mathbb{R}$, 
    $z \in \mathbb{C}$ and $\tilde{{\bf I}} \in \mathbb{R}^{m \times n}$ (the identity 
    matrix with ones on the main diagonal and zeros elsewhere), then the $\epsilon$-pseudospectra
    $\Lambda_{\epsilon} = \{z \in \mathbb{C}: \sigma_{\min}(z\tilde{{\bf I}}-{\bf V})\leq \epsilon \}$.
\end{definition}

Definitions~\ref{def:psd1} and~\ref{def:psd2} are equivalent~\cite{wright:2002}, and both 
account for the computation of $\Lambda_{\epsilon}$ of rectangular matrices of dimension
$m \times n$, where $m \geq n$. The algorithm given in~\cite{wright:2002} for 
the numerical computation of the set $\Lambda_{\epsilon}$ can be applied on matrices of 
dimension $m \times n$, where either $m \geq 2n$ or $m \leq 2n$. 

\subsection{Error and Accuracy Table}
\label{appendix:table_err_acc}

Here we provide the minimum test error and maximum test accuracy for each of 
the experiments (XOR, MNIST and eMNIST). We compare in the following table 
our CHL and rCHL implementations against each other and against the BP and 
the FDA.

\begin{table}[htpb!]
\centering
    \begin{tabular}{cccc}
 \diagbox[width=10em]{Algorithm}{Experiment}  & XOR & MNIST & eMNIST \\ 
\hline
    BP              & $0.003/1$    &  $0.0087/0.962$     & $0.0116/0.8935$       \\  
    FDA             & $0.0009/1$   &  $0.0066/0.969$     & $0.0087/0.9035$       \\ 
    CHL             & $\expnumber{2.503}{-8}/1$   &  $0.0038/0.977$  &  $0.0061/0.899$      \\ 
    rCHL 			& ${\bf \expnumber{1.117}{-6}/1}$    & ${\bf 0.0044/0.975}$  &  ${\bf0.0073/0.8461}$     \\ 
\end{tabular}
\label{table:0}
    \caption{{\bfseries \sffamily Mean squared error (MSE) and Accuracy.} The test 
    MSE and accuracy of three experiments, XOR, MNIST, and eMNIST are given in this
    table. BP--Backpropagation, FDA--Feedback Alignment, CHL-- Contrastive Hebbian
    Learning, and rCHL--random Contrastive Hebbian Learning.}
\end{table}

\subsection{Abbreviations and Notation Tables}

\begin{table}[htpb!]
\centering
\begin{tabular}{l|l}
Abbreviation & Description \\
\hline \\
STDP	& Spike-timing Dependent Plasticity \\
BP		& Back-propagation \\
FDA		& Feedback Alignment \\
CHL		& Contrastive Hebbian Learning \\
rCHL 	& random Contrastive Hebbian Learning \\
MSE		& Mean Squared Error 
\end{tabular}
\label{table:abbrev}
\caption{{\bfseries \sffamily Abbreviations}}
\end{table}

\begin{table}[htpb!]
\centering
\begin{tabular}{l|l}
Symbols & Description \\
\hline \\
$L$			& Total number of layers \\
$k$			& Index of layer \\
${\bf x}_k$	& Neural state at layer $k$ \\
$\check{{\bf x}}_k$ & Neural state at layer $k$ in the free phase \\
$\hat{{\bf x}}_k$ & Neural state at layer $k$ in the clamped phase \\
${\bf W}_k$ & Synaptic matrix (connects layer $k-1$ with $k$) \\
${\bf G}_k$	& Random feedback matrix (connects layer $k-1$ with $k$ \\
${\bf b}_k$ & Bias for layer $k$ \\
$f_k$		& Non-linear function (transfer function) \\
$\eta$		& Learning rate \\
$\gamma$	& Feedback gain \\
$\otimes$	& Tensor product \\
$dt$		& Forward Euler time-step \\
$t_f$		& Forward Euler total integration time \\
$\mathcal{I}$	& Input data set \\
$\mathcal{T}$ 	& Target set \\
$\Lambda_{\epsilon}$	& Set of pseudospectra \\
$\lambda$		& Eigenvalue \\
$\mathcal{U}$	& Uniform distribution \\
$\ell$			& Length of uniform's distribution interval \\
$\mathcal{N}$	& Normal distribution \\
$\sigma$		& Normal distribution's variance \\
$\oplus$		& Exclusive Or (XOR)
\end{tabular}
\label{table:notation}
\caption{{\bfseries \sffamily Notation}}
\end{table}

\subsection{Simulation and Platform Details}
\label{appendix:sim_details}

All serial simulations were run on a Dell OptiPlex $7040$ with $16$GB physical memory, 
and a $6$th generation Intel i$7$ processor (Intel(R) Core(TM) i7-6700 CPU @ 3.40GHz)
running Arch Linux ($4.11.9-1$-ARCH $\times 86\_64$ GNU/Linux). 
The source code for CHL and rCHL were written in the C programming 
language~\cite{kernighan:2006} (gcc (GCC) $7.1.1 20170630$).
In all C simulations, we used the random number generator provided by
\cite{oneill:2014}\footnote{http://www.pcg-random.org/}.
The backpropagation and feedback alignment algorithms were written in Python
using Tensorflow \cite{abadi:2016}, and ran on an Nvidia GeForce GTX Titan X
with $12\, \mathrm{GB}$ memory. The
source code is distributed under the GNU General Public License, and can be found at: [LINK].
All simulation parameters are provided in the Results section and in the Supplementary 
Information (SI).

\subsection{Simulation Parameters}
\label{appendix:sim_params}

\begin{table}[htpb!]
\centering
\resizebox{\columnwidth}{!}{%
\begin{tabular}{l*{7}{c}}
Experiment       & $\eta$ & $\gamma$ & Epochs & Layout & ${\bf G}_k$ & ${\bf W}_k^0$ & ${\bf b}_k^0$  \\
\hline
XOR    		     & $0.1$ & $0.05$ & $5,000$ & $(2,2,1)$ & $\mathcal{U}(-0.2, 0.2)$ & $\mathcal{U}(0, 1)$ & $-$ \\
MNIST            & $0.1$ & $0.05$ & $20$ & $(784,128,64,10)$ & $\mathcal{U}(-0.6, 0.6)$ & $\mathcal{U}(-0.5, 0.5)$ 
&  $-$ \\
eMNIST 			 & $0.1$ & $0.1$ & $40$ & $(784,256,128,26)$ & $\mathcal{U}(-0.3, 0.3)$ & $\mathcal{U}(-0.5, 0.5)$  
& $\mathcal{U}(-0.2, 0.2)$ \\
Autoencoder      & $0.05$ & $0.05$ & $20$ & $(784,36,784)$ & $\mathcal{U}(-1, 1)$ & $\mathcal{U}(-0.5, 0.5)$ & $-$ 
\end{tabular}
}
\label{table:1}
\caption{{\bfseries \sffamily Simulation Parameters.} In the experiments, we integrated
the dynamics for $t_f = 30\, \mathrm{ms}$ using the forward Euler method with time-step $dt = 0.08$.
$\eta$ is the learning rate and $\gamma$ is the feedback gain.}
\end{table}

\section*{Supplementary Information}

\subsection*{Backpropagation and Feedback Alignment on XOR}

\subsection*{Backpropagation and Feedback Alignment on MNIST}

\subsection*{Backpropagation and Feedback Alignment on eMNIST}

\bibliographystyle{plain}
\bibliography{biblio}

\begin{thebibliography}{10}

\bibitem{abadi:2016}
Mart{\'\i}n Abadi, Paul Barham, Jianmin Chen, Zhifeng Chen, Andy Davis, Jeffrey
  Dean, Matthieu Devin, Sanjay Ghemawat, Geoffrey Irving, Michael Isard, et~al.
\newblock Tensorflow: A system for large-scale machine learning.
\newblock In {\em OSDI}, volume~16, pages 265--283, 2016.

\bibitem{abbott:2000}
Larry~F Abbott and Sacha~B Nelson.
\newblock Synaptic plasticity: taming the beast.
\newblock {\em Nature neuroscience}, 3(11s):1178, 2000.

\bibitem{baldi:1991}
Pierre Baldi and Fernando Pineda.
\newblock Contrastive learning and neural oscillations.
\newblock {\em Neural Computation}, 3(4):526--545, 1991.

\bibitem{bi:1998}
Guo-qiang Bi and Mu-ming Poo.
\newblock Synaptic modifications in cultured hippocampal neurons: dependence on
  spike timing, synaptic strength, and postsynaptic cell type.
\newblock {\em Journal of neuroscience}, 18(24):10464--10472, 1998.

\bibitem{cohen:2017}
Gregory Cohen, Saeed Afshar, Jonathan Tapson, and Andr{\'e} van Schaik.
\newblock Emnist: an extension of mnist to handwritten letters.
\newblock {\em arXiv preprint arXiv:1702.05373}, 2017.

\bibitem{cooper:2005}
Steven~J Cooper.
\newblock Donald o. hebb's synapse and learning rule: a history and commentary.
\newblock {\em Neuroscience \& Biobehavioral Reviews}, 28(8):851--874, 2005.

\bibitem{deng:2012}
Li~Deng.
\newblock The mnist database of handwritten digit images for machine learning
  research [best of the web].
\newblock {\em IEEE Signal Processing Magazine}, 29(6):141--142, 2012.

\bibitem{dreyfus:1962}
Stuart Dreyfus.
\newblock The numerical solution of variational problems.
\newblock {\em Journal of Mathematical Analysis and Applications}, 5(1):30--45,
  1962.

\bibitem{elde:2012}
Nels~C Elde, Stephanie~J Child, Michael~T Eickbush, Jacob~O Kitzman, Kelsey~S
  Rogers, Jay Shendure, Adam~P Geballe, and Harmit~S Malik.
\newblock Poxviruses deploy genomic accordions to adapt rapidly against host
  antiviral defenses.
\newblock {\em Cell}, 150(4):831--841, 2012.

\bibitem{gerstner:2002}
Wulfram Gerstner and Werner~M Kistler.
\newblock {\em Spiking neuron models: Single neurons, populations, plasticity}.
\newblock Cambridge university press, 2002.

\bibitem{glasser:2016}
Matthew~F Glasser, Timothy~S Coalson, Emma~C Robinson, Carl~D Hacker, John
  Harwell, Essa Yacoub, Kamil Ugurbil, Jesper Andersson, Christian~F Beckmann,
  Mark Jenkinson, et~al.
\newblock A multi-modal parcellation of human cerebral cortex.
\newblock {\em Nature}, 536(7615):171--178, 2016.

\bibitem{goodfellow:2016}
Ian Goodfellow, Yoshua Bengio, and Aaron Courville.
\newblock {\em Deep Learning}.
\newblock MIT Press, 2016.

\bibitem{harris:2013}
Kenneth~D Harris and Thomas~D Mrsic-Flogel.
\newblock Cortical connectivity and sensory coding.
\newblock {\em Nature}, 503(7474):51, 2013.

\bibitem{hebb:2005}
Donald~Olding Hebb.
\newblock {\em The organization of behavior: A neuropsychological theory}.
\newblock Psychology Press, 2005.

\bibitem{hinton:2002}
Geoffrey~E Hinton.
\newblock Training products of experts by minimizing contrastive divergence.
\newblock {\em Neural computation}, 14(8):1771--1800, 2002.

\bibitem{hinton:1988}
Geoffrey~E Hinton and James~L McClelland.
\newblock Learning representations by recirculation.
\newblock In {\em Neural information processing systems}, pages 358--366, 1988.

\bibitem{kandel:2000}
Eric~R Kandel, James~H Schwartz, Thomas~M Jessell, Steven~A Siegelbaum, A~James
  Hudspeth, et~al.
\newblock {\em Principles of neural science}, volume~4.
\newblock McGraw-hill New York, 2000.

\bibitem{kernighan:2006}
Brian~W Kernighan and Dennis~M Ritchie.
\newblock {\em The C programming language}.
\newblock 2006.

\bibitem{le:1986}
Yann Le~Cun.
\newblock Learning process in an asymmetric threshold network.
\newblock In {\em Disordered systems and biological organization}, pages
  233--240. Springer, 1986.

\bibitem{lecun:2015}
Yann LeCun, Yoshua Bengio, and Geoffrey Hinton.
\newblock Deep learning.
\newblock {\em nature}, 521(7553):436, 2015.

\bibitem{lee:2015}
Dong-Hyun Lee, Saizheng Zhang, Asja Fischer, and Yoshua Bengio.
\newblock Difference target propagation.
\newblock In {\em Joint european conference on machine learning and knowledge
  discovery in databases}, pages 498--515. Springer, 2015.

\bibitem{lillicrap:2016}
Timothy~P Lillicrap, Daniel Cownden, Douglas~B Tweed, and Colin~J Akerman.
\newblock Random synaptic feedback weights support error backpropagation for
  deep learning.
\newblock {\em Nature communications}, 7, 2016.

\bibitem{mackay:2003}
David~JC MacKay.
\newblock {\em Information theory, inference, and learning algorithms},
  volume~7.
\newblock Citeseer, 2003.

\bibitem{macknik:2007}
Stephen~L Macknik and Susana Martinez-Conde.
\newblock The role of feedback in visual masking and visual processing.
\newblock {\em Advances in cognitive psychology}, 2007.

\bibitem{macknik:2009}
Stephen~L Macknik and Susana Martinez-Conde.
\newblock The role of feedback in visual attention and awareness.
\newblock {\em Cognitive Neurosciences}, 1, 2009.

\bibitem{markov:2012}
Nikola~T Markov, MM~Ercsey-Ravasz, AR~Ribeiro~Gomes, Camille Lamy, Loic Magrou,
  Julien Vezoli, P~Misery, A~Falchier, R~Quilodran, MA~Gariel, et~al.
\newblock A weighted and directed interareal connectivity matrix for macaque
  cerebral cortex.
\newblock {\em Cerebral cortex}, 24(1):17--36, 2012.

\bibitem{markov:2014}
Nikola~T Markov, Julien Vezoli, Pascal Chameau, Arnaud Falchier, Ren{\'e}
  Quilodran, Cyril Huissoud, Camille Lamy, Pierre Misery, Pascale Giroud,
  Shimon Ullman, et~al.
\newblock Anatomy of hierarchy: feedforward and feedback pathways in macaque
  visual cortex.
\newblock {\em Journal of Comparative Neurology}, 522(1):225--259, 2014.

\bibitem{markram:1997}
Henry Markram, Joachim L{\"u}bke, Michael Frotscher, and Bert Sakmann.
\newblock Regulation of synaptic efficacy by coincidence of postsynaptic aps
  and epsps.
\newblock {\em Science}, 275(5297):213--215, 1997.

\bibitem{may:1976}
Robert~M May.
\newblock Simple mathematical models with very complicated dynamics.
\newblock {\em Nature}, 261(5560):459, 1976.

\bibitem{minsky:1961}
Marvin Minsky.
\newblock Steps toward artificial intelligence.
\newblock {\em Proceedings of the IRE}, 49(1):8--30, 1961.

\bibitem{movellan:1991}
Javier~R Movellan.
\newblock Contrastive hebbian learning in the continuous hopfield model.
\newblock In {\em Connectionist models: Proceedings of the 1990 summer school},
  pages 10--17, 1991.

\bibitem{neftci:2014}
Emre Neftci, Srinjoy Das, Bruno Pedroni, Kenneth Kreutz-Delgado, and Gert
  Cauwenberghs.
\newblock Event-driven contrastive divergence for spiking neuromorphic systems.
\newblock {\em Frontiers in neuroscience}, 7:272, 2014.

\bibitem{neftci:2017}
Emre~O Neftci, Charles Augustine, Somnath Paul, and Georgios Detorakis.
\newblock Event-driven random back-propagation: Enabling neuromorphic deep
  learning machines.
\newblock {\em Frontiers in neuroscience}, 11, 2017.

\bibitem{neftci:2016}
Emre~O Neftci, Bruno~U Pedroni, Siddharth Joshi, Maruan Al-Shedivat, and Gert
  Cauwenberghs.
\newblock Stochastic synapses enable efficient brain-inspired learning
  machines.
\newblock {\em Frontiers in neuroscience}, 10:241, 2016.

\bibitem{nokland:2016}
Arild N{\o}kland.
\newblock Direct feedback alignment provides learning in deep neural networks.
\newblock In {\em Advances in Neural Information Processing Systems}, pages
  1037--1045, 2016.

\bibitem{oh:2014}
Seung~Wook Oh, Julie~A Harris, Lydia Ng, Brent Winslow, Nicholas Cain, Stefan
  Mihalas, Quanxin Wang, Chris Lau, Leonard Kuan, Alex~M Henry, et~al.
\newblock A mesoscale connectome of the mouse brain.
\newblock {\em Nature}, 508(7495):207, 2014.

\bibitem{oneill:2014}
Melissa~E. O'Neill.
\newblock Pcg: A family of simple fast space-efficient statistically good
  algorithms for random number generation.
\newblock Technical Report HMC-CS-2014-0905, Harvey Mudd College, Claremont,
  CA, September 2014.

\bibitem{oreily:1996}
Randall~C O'Reilly.
\newblock Biologically plausible error-driven learning using local activation
  differences: The generalized recirculation algorithm.
\newblock {\em Neural computation}, 8(5):895--938, 1996.

\bibitem{petrantonakis:2014}
Panagiotis~C Petrantonakis and Panayiota Poirazi.
\newblock A compressed sensing perspective of hippocampal function.
\newblock {\em Frontiers in systems neuroscience}, 8, 2014.

\bibitem{pickel:2013}
Virginia~M Pickel and Menahem Segal.
\newblock {\em The synapse: structure and function}.
\newblock Elsevier, 2013.

\bibitem{rougier:2011}
Nicolas~P Rougier and Axel Hutt.
\newblock Synchronous and asynchronous evaluation of dynamic neural fields.
\newblock {\em Journal of Difference Equations and Applications},
  17(8):1119--1133, 2011.

\bibitem{rumelhart:1988}
David~E Rumelhart, Geoffrey~E Hinton, Ronald~J Williams, et~al.
\newblock Learning representations by back-propagating errors.
\newblock {\em Cognitive modeling}, 5(3):1, 1988.

\bibitem{shou:2010}
Tian-De Shou.
\newblock The functional roles of feedback projections in the visual system.
\newblock {\em Neuroscience bulletin}, 26(5):401--410, 2010.

\bibitem{taouali:2009}
Wahiba Taouali, Fr{\'e}d{\'e}ric Alexandre, Axel Hutt, and Nicolas~P Rougier.
\newblock Asynchronous evaluation as an efficient and natural way to compute
  neural networks.
\newblock In {\em AIP Conference Proceedings}, volume 1168, pages 554--558.
  AIP, 2009.

\bibitem{thomson:2003}
Alex~M Thomson and A~Peter Bannister.
\newblock Interlaminar connections in the neocortex.
\newblock {\em Cerebral cortex}, 13(1):5--14, 2003.

\bibitem{trefethen:1991}
Lloyd~N Trefethen.
\newblock Pseudospectra of matrices.
\newblock {\em Numerical analysis}, 91:234--266, 1991.

\bibitem{trefethen:2005}
Lloyd~N Trefethen and Mark Embree.
\newblock {\em Spectra and pseudospectra: the behavior of nonnormal matrices
  and operators}.
\newblock Princeton University Press, 2005.

\bibitem{vu:2014}
Van~H Vu.
\newblock {\em Modern Aspects of Random Matrix Theory}, volume~72.
\newblock American Mathematical Society, 2014.

\bibitem{wright:2002}
Thomas~G Wright and Lloyd~N Trefethen.
\newblock Pseudospectra of rectangular matrices.
\newblock {\em IMA Journal of Numerical Analysis}, 22(4):501--519, 2002.

\bibitem{xie:2003}
Xiaohui Xie and H~Sebastian Seung.
\newblock Equivalence of backpropagation and contrastive hebbian learning in a
  layered network.
\newblock {\em Neural computation}, 15(2):441--454, 2003.

\bibitem{zhang:1998}
Li~I Zhang, Huizhong~W Tao, Christine~E Holt, William~A Harris, and Mu-ming
  Poo.
\newblock A critical window for cooperation and competition among developing
  retinotectal synapses.
\newblock {\em Nature}, 395(6697):37, 1998.

\end{thebibliography}

\end{document}